\documentclass[10pt,twocolumn,letterpaper]{article}

\usepackage[pagenumbers]{cvpr} 

\usepackage{graphicx}
\usepackage{amsmath}
\usepackage{amssymb}
\usepackage{booktabs}

\usepackage{cite}
\usepackage{amssymb}
\usepackage{mathtools}
\usepackage{mathrsfs}
\usepackage{url}
\usepackage{amsfonts}
\usepackage{paralist}
\usepackage{afterpage}
\usepackage{multirow}
\usepackage{float}
\usepackage[subrefformat=parens]{subcaption}
\usepackage{xcolor}
\usepackage[accsupp]{axessibility}

\newcommand{\bm}[1]{\mbox{\boldmath ${#1}$}}
\usepackage{lmodern} 
\usepackage{scalerel}
\newcommand\wh[1]{\hstretch{2}{\hat{\hstretch{.5}{#1}}}}

\usepackage[pagebackref,breaklinks,colorlinks]{hyperref}

\usepackage[capitalize]{cleveref}
\crefname{section}{Sec.}{Secs.}
\Crefname{section}{Section}{Sections}
\Crefname{table}{Table}{Tables}
\crefname{table}{Tab.}{Tabs.}

\def\cvprPaperID{8022} 
\def\confName{CVPR}
\def\confYear{2022}

\begin{document}

\title{MotionAug: Augmentation with Physical Correction\\ for Human Motion Prediction}

\author{Takahiro Maeda \qquad Norimichi Ukita\\
Toyota Technological Institute, Japan\\
{\tt\small \{sd21601, ukita\}@toyota-ti.ac.jp}}

\maketitle

\begin{figure*}[t]
    \centering
	\includegraphics[width=0.95\textwidth]{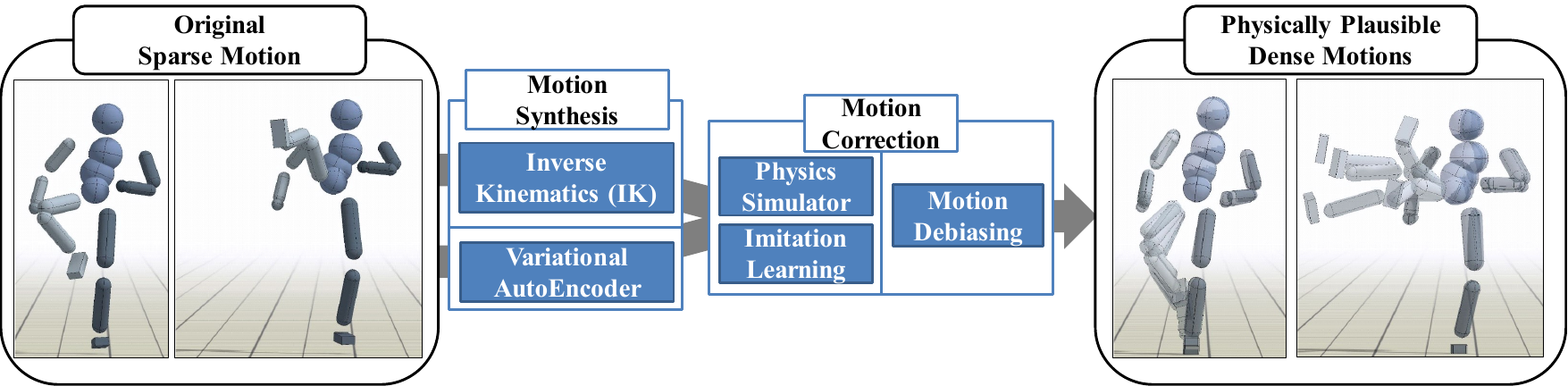}
	\caption{\textbf{Overview of our proposed motion data augmentation.} Original motions are augmented independently using our VAE- and IK-based syntheses. We also propose (i) motion correction where the synthesized motions are modified to be physically-plausible using imitation learning, a physics simulator, and (ii) subsequent motion debiasing.}
	\label{fig:overview}
	\vspace{-3mm}
\end{figure*}

\begin{abstract}
This paper presents a motion data augmentation scheme incorporating motion synthesis encouraging diversity and motion correction imposing physical plausibility.
This motion synthesis consists of our modified Variational AutoEncoder (VAE) and Inverse Kinematics (IK).
In this VAE, our proposed sampling-near-samples method generates various valid motions even with insufficient training motion data.
Our IK-based motion synthesis method allows us to generate a variety of motions semi-automatically.
Since these two schemes generate unrealistic artifacts in the synthesized motions, our motion correction rectifies them.
This motion correction scheme consists of imitation learning with physics simulation and subsequent motion debiasing.
For this imitation learning, we propose the PD-residual force that significantly accelerates the training process.
Furthermore, our motion debiasing successfully offsets the motion bias induced by imitation learning to maximize the effect of augmentation.
As a result, our method outperforms previous noise-based motion augmentation methods by a large margin on both Recurrent Neural Network-based and Graph Convolutional Network-based human motion prediction models.
The code is available at {\rm \url{https://github.com/meaten/MotionAug}}.
\vspace{-5mm}
\end{abstract}

\section{Introduction}

Human motion prediction, which forecasts future body poses based on past poses, is a crucial technique for human-robot interaction~\cite{liu2017human, kim2017anticipatory, peternel2017towards, lorenzini2018synergistic, tortora2019fast}, autonomous driving~\cite{ma2019trafficpredict}, VR/AR applications~\cite{hou2020motion}, performance capture~\cite{DBLP:conf/iccv/UkitaHK09,DBLP:journals/cviu/UkitaK12,martinez2017human,DBLP:conf/mva/MatsumotoSMMMMU19,DBLP:conf/cvpr/CoronaPAM20}, etc.
However, these applications are limited because of the lack of training motion data, which results in low prediction accuracy.
This data insufficiency is caused by the enormous cost of motion data acquisition, such as motion capture equipment, recordings, post-processing, and denoising.

Such data insufficiency can be alleviated by Data Augmentation (DA), for example, image recognition~\cite{tanner1987calculation,krizhevsky2012imagenet}.
%
Compared with image data augmentation, however, DA for motion data is hard to address because simple numerical transformations (\eg, additive noise) may generate physically-implausible motions such as too high velocities or floating motions.

This paper presents a novel motion data augmentation approach, including motion synthesis and motion correction.
Our motion synthesis uses Variational AutoEncoder (VAE)~\cite{DBLP:journals/corr/KingmaW13} to exploit a training data distribution and Inverse Kinematics (IK) to exploit human knowledge, shown in ``Motion Synthesis'' in Fig.~\ref{fig:overview}. 
Although most of our synthesized motions are physically-plausible, we observed some of them have unrealistic artifacts, which lead to the low accuracy of human motion prediction.
These artifacts are corrected with our proposed motion correction method.
This correction method uses (i) imitation learning with physics simulation to rectify these artifacts and (ii) subsequent motion debiasing to offset biases imposed by the mismatch between the bodies of a human and the character during imitation learning (``Motion Correction'' in Fig.~\ref{fig:overview}).
Our contributions for motion diversity and physical plausibility are as follows:
\begin{enumerate}
    \item {\bf VAE-based human-motion synthesis:} Our generative model with adversarial training in sequencewise and framewise, and sampling-near-samples can generate plausible motions even with insufficient motion data.
    \item {\bf IK-based human-motion synthesis:} Compared with annotating IK target points in all frames as the standard IK does, our method requires less effort because only a target sampling space for a keyframe is manually given.
    \item {\bf PD-residual force:} We propose the PD-residual force that accelerates the training of imitation learning in a physics simulator to rectify the physical implausibility of synthesized motions.
    \item {\bf Motion debiasing:} Our motion debiasing successfully offsets the motion bias induced by the imitation learning to maximize the effect of our data augmentation.
\end{enumerate}

\section{Related Work}

\subsection{Human Motion Prediction and Augmentation}

From the releases of large-scale motion capture sequence datasets~\cite{de2009guide, h36m_pami}, 
many deep learning-based human motion prediction methods were proposed.
Most approaches~\cite{fragkiadaki2015recurrent, jain2016structural, martinez2017human, DBLP:conf/iclr/ZhouLXHH018, ghosh2017learning, DBLP:conf/bmvc/PavlloGA18, DBLP:conf/wacv/ChiuAWHN19, gopalakrishnan2019neural, wang2019imitation} are built upon Recurrent Neural Networks (RNN) to model sequence-to-sequence relationships between past and future motions. 
Recently, Graph Convolutional Network (GCN)-based approaches~\cite{DBLP:conf/iccv/MaoLSL19} achieved a better performance than RNN-based models by encoding motions with Discrete Cosine Transform.
Along with improving model architectures, the stochasticity of human motion is addressed by using generative models such as Generative Adversarial Networks (GAN)~\cite{DBLP:conf/cvpr/BarsoumKL18, kundu2019bihmp}, VAE~\cite{yan2018mt, DBLP:conf/cvpr/AliakbarianSSPG20}, and Flow-based models~\cite{yuan2020dlow}.

These approaches assume a large-scale motion dataset that is too expensive to obtain in real-world tasks.
Despite this difficulty,
motion data augmentation approaches are almost ignored.
Fragkiadaki \etal~\cite{fragkiadaki2015recurrent} proposed corrupting input motions with zero-mean Gaussian noise for motion data augmentation. 
While this simple additive noise improves the variety of input motions, the augmented motions might lose motion contexts and defy the laws of physics. 


\subsection{Data augmentation with generative models}

In image classification tasks, generative models such as GAN are used for data augmentation by generating within-class images~\cite{tran2017bayesian, huang2018auggan, choi2019self}.
This approach is applicable to other tasks, including image segmentation~\cite{sandfort2019data} and person re-identification~\cite{zhang2020pac}.
However, generative models for human motion might synthesize several kinds of physically-implausible motions because it is difficult to learn such physical plausibility from a limited number of training motion data, especially in data insufficient settings.

\subsection{Inverse Kinematics (IK)}

IK modifies the pose of a whole body so that key points in the body reach their target positions.
IK can also modify a motion by providing the target positions in all frames of the motion~\cite{gleicher2001motion}. 
Although IK can significantly modify each pose and potentially be helpful for motion augmentation, it is impractical to manually annotate the target positions in all frames included in a training dataset for augmenting all motions~\cite{DBLP:journals/jvca/CarvalhoBT07, ho2009character}.

\subsection{Motion Synthesis with Physics Simulation}

Motion prediction models should be trained on physically-plausible motions for better accuracy and reliability.
Therefore, physics simulation might improve the quality of augmented motions.
Recent deep reinforcement learning enables a physically-simulated character to imitate various motions~\cite{peng2018deepmimic, DBLP:journals/tog/BergaminCHF19, lee2019scalable, yuan2020residual, yuan2021simpoe}.
However, these methods often require more than one day to converge for imitating only one motion.
To incorporate this, we need to reduce the vast computational cost to augment a large number of motions.

\section{Proposed Motion Augmentation}

We propose two independent motion synthesis approaches with VAE and IK, described in Secs.~\ref{subsec:VAE} and \ref{subsec:DA_with_IK}, respectively.
Furthermore, a method for motion correction is also proposed for rectifying the artifacts of synthesized motions in Sec.~\ref{subsec:imitation_learning}.
Finally, we propose motion debiasing to offset the bias imposed by dynamic mismatch, as  presented in Sec.~\ref{subsec:debiasing}.

\begin{figure*}
    \centering
    \includegraphics[width=0.9\textwidth]{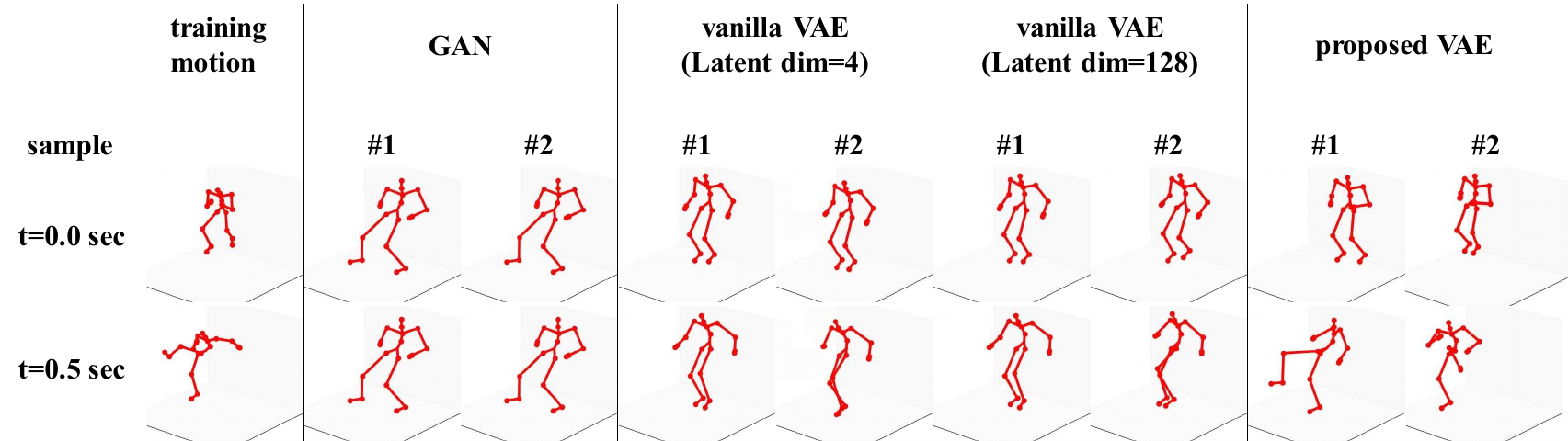}
    \caption{\textbf{Synthesized motions from GAN, vanilla VAE, and our VAE.} Despite the dynamic training motions, GAN produces static motions due to data insufficiency. The vanilla VAE produces non-diverse motions regardless of the dimension of the latent space. Our VAE with adversarial training and sampling-near-samples successfully synthesizes dynamic motions different from training motions. }
    \label{fig:vae_example}
    \vspace{-3mm}
\end{figure*}

\begin{figure}[t]
    \centering
    \includegraphics[width=0.85\linewidth]{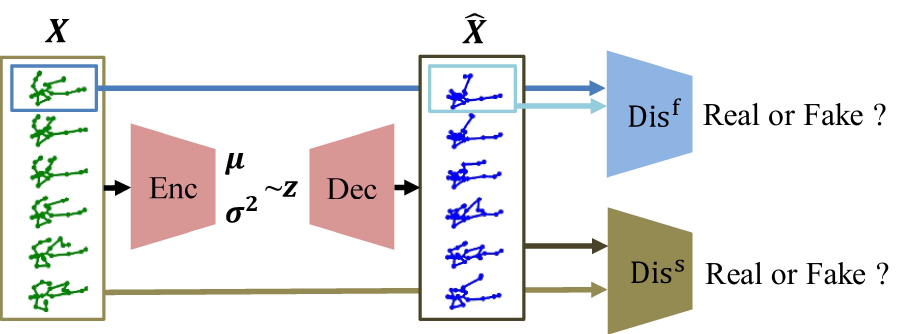}
    \caption{\textbf{Proposed VAE-based network  with adversarial training.}
    The synthesized motions and training motions are discriminated framewise and sequencewise.}
    \label{fig:vae_network}
    \vspace{-3mm}
\end{figure}

\begin{figure*}[t]
    \centering
    \begin{minipage}[t]{0.305\textwidth}
    \centering
    \includegraphics[width=\textwidth]{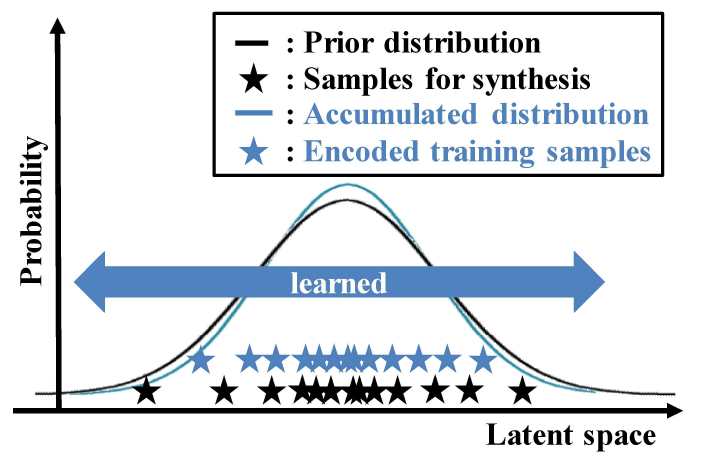}
    \subcaption{VAE with enough data}\label{fig/sns_enough}
  \end{minipage}
  \begin{minipage}[t]{0.32\textwidth}
    \centering
    \includegraphics[width=\textwidth]{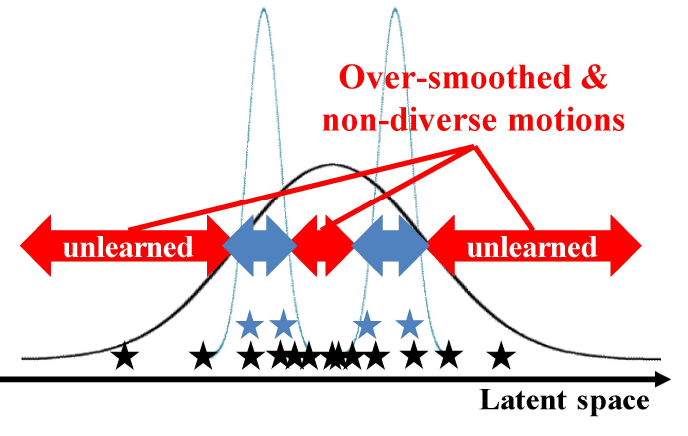}
    \subcaption{VAE with insufficient data}\label{fig/sns_insufficient}
  \end{minipage}
  \begin{minipage}[t]{0.32\textwidth}
    \centering
    \includegraphics[width=\textwidth]{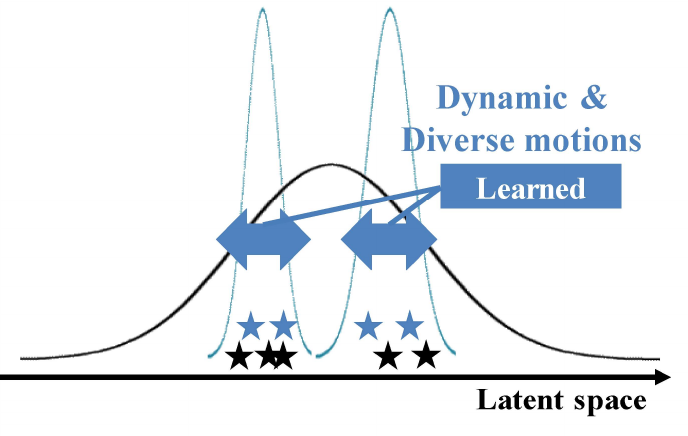}
    \subcaption{VAE using sampling-near-samples with insufficient data }\label{fig/sns_sns}
  \end{minipage}
    \caption{\textbf{Our sampling-near-samples for insufficient data.} The prior distribution matches the accumulated distribution that aggregates the distributions of encoded training data with enough data (a). Synthesized motions are mostly sampled from learned regions (\textcolor{blue}{blue arrows}). With insufficient data (b), the accumulated distribution (learned regions) gets sparse, and the prior distribution often samples from the unlearned regions (\textcolor{red}{red arrows}), which leads to over-smoothed and non-diverse motions. We propose sampling-near-samples (c), which samples only from learned regions by sampling latent representations using the clusters of training motions.}
    \label{fig:sampling_near_samples}
    \vspace{-3mm}
\end{figure*}

\subsection{DA with VAE}
\label{subsec:VAE}

Although GAN is widely used as a generative model, we found that, for motion synthesis, GAN often produces only static motions where all poses are almost identical due to data insufficiency and training instability of GAN (i.e., mode collapse).
Instead, we propose a VAE-based model that is free from these problems.
Our VAE-based model described below successfully generates various motions despite insufficient data using adversarial training and sampling-near-samples, as shown in Fig. \ref{fig:vae_example}.

\noindent \textbf{Adversarial training:}
Our proposed network is shown in Fig.~\ref{fig:vae_network}.
The encoder produces a mean $\bm{\mu}$ and a variance $\bm{\sigma}^2$ in the latent space from an input motion $\bm{X} = \{\bm{x}_1, \bm{x}_2, \ldots, \bm{x}_T\}$ where each $\bm{x}_{t}$ denotes a pose vector in $t$-th frame.
The latent representation $\bm{z}$ is sampled from the normal distribution $\mathcal{N}(\bm{\mu}, \bm{\sigma}^2)$.
The decoder reconstructs a motion $\wh{\bm{X}}$ from $\bm{z}$.
Frame-wise and sequence-wise discriminators (denoted by $Dis^{f}$ and $Dis^{s}$, respectively, in Fig.~\ref{fig:vae_network}) discriminate $\bm{X}$ from $\wh{\bm{X}}$ for improving $\wh{\bm{X}}$ in terms of the fidelity of poses and motion dynamics.
We validated that this VAE with adversarial training can suppress mode collapse and generate more realistic motions than the vanilla VAE.

\noindent \textbf{Sampling-near-samples in the latent space:}
In the inference of the vanilla VAE, a latent representation $\bm{z}$ is sampled from a normal distribution with zero mean and unit variance $\mathcal{N}(\bm{0}, \bm{I})$.
This normal distribution should be represented well by all training samples in the latent space for the better quality of synthesized motions, as shown in Fig.~\ref{fig/sns_enough}.

However, the dimension of the latent space should not be set too low so that insufficient data could cover the whole normal distribution because too low-dimensional latent space has a too narrow bottleneck and generates inaccurate motions that lack motion details.
On the other hand, a high-dimensional representation leads to the sparsity of training data, making it difficult to sample realistic data from the learned regions, as shown in Fig.~\ref{fig/sns_insufficient}.
Therefore, we have a tradeoff between motion details and sampling easiness.


To solve this tradeoff, we propose a novel sampling method robust to sparsity, specifically sampling $\bm{z}$ from only learned regions that are appropriately represented by training data in the latent space, as shown in Fig.~\ref{fig/sns_sns}.
In this method, each motion in the training data is encoded into mean $\bm{\mu}$ and variance $\bm{\sigma}^2$.
We apply k-means clustering to all training motions based on $\bm{\mu}$ to make $n_c$ clusters.
Given $\overline{\bm{\mu}}$ and $\overline{\bm{\sigma^2}}$ that are respectively the mean of $\bm{\mu}$ and the mean of $\bm{\sigma}^2$ over randomly-sampled $n_s$ training motions from each cluster,
the latent representation $\bm{z}$ is drawn from $\mathcal{N}(\overline{\bm{\mu}}, \overline{\bm{\sigma}^2})$, and
$\bm{z}$ is fed into the decoder for generating $\bm{X}_{\rm aug}$, as expressed by $\bm{X}_{\rm aug} = \textrm{Dec}(\bm{z}) \label{eq:dec}$ and $\bm{z} \sim \mathcal{N}(\overline{\bm{\mu}}, \overline{\bm{\sigma}^2}) \label{eq:z}$.
We sampled motion subsets from each cluster for efficiency. 
\subsection{DA with IK}
\label{subsec:DA_with_IK}

\begin{figure}[t]
    \centering
    \includegraphics[width=0.8\linewidth]{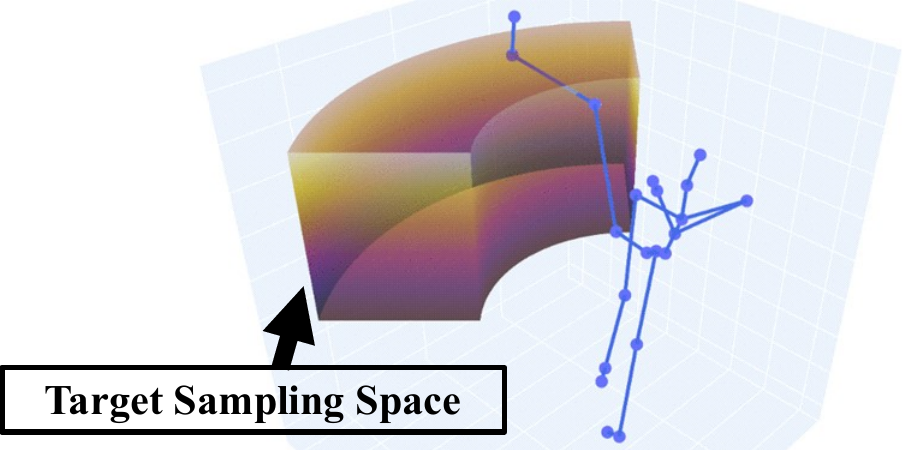}
    \caption{\textbf{Target sampling space for the action class \textit{kick}.} The target space is a fan-shaped one that a foot end-effector may reach. We uniformly sample targets $\bm{p}^{\rm sample}_{t_{\rm key}}$ from this space for the keyframe $t_{\rm key}$.}
    \label{fig:kick_space}
    \vspace{-3mm}
\end{figure}

\begin{figure*}[t]
    \centering
    \includegraphics[width=0.85\textwidth]{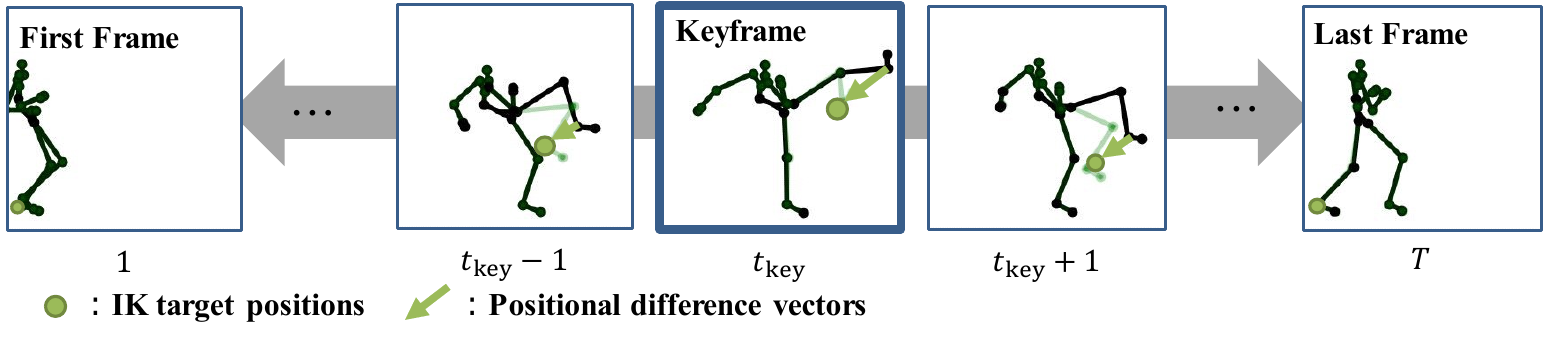}
    \caption{\textbf{Overview of our sequential IK scheme.}
    Given an IK target and the body pose in the keyframe, body poses in all other frames are automatically calculated by IK. IK target positions are automatically determined by propagating the positional difference on the keyframe in a linearly decreasing manner.}
    \label{fig:target_propagation}
    \vspace{-3mm}
\end{figure*}

\begin{figure}[t]
\centering
\begin{minipage}[t]{0.3\linewidth}
    \includegraphics[width=\textwidth]{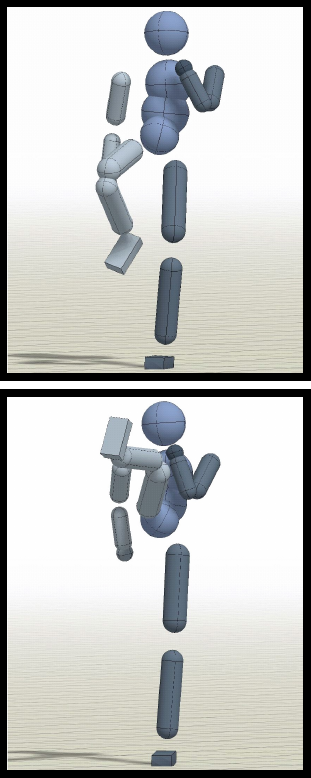}
    \subcaption{original motion}\label{fig:correct_original}
  \end{minipage}
  \begin{minipage}[t]{0.3\linewidth}
    \includegraphics[width=\textwidth]{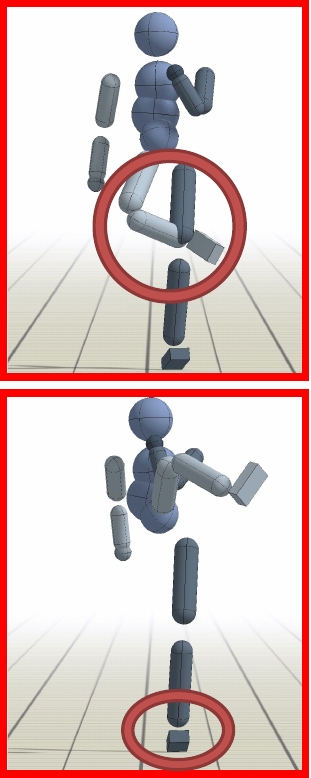}
    \subcaption{synthe'd motion}\label{fig:correct_synthesized}
  \end{minipage}
  \begin{minipage}[t]{0.303\linewidth}
    \includegraphics[width=\textwidth]{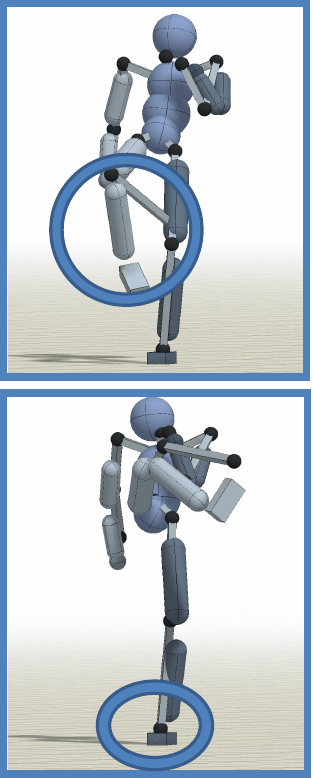}
    \subcaption{rectified motion}\label{fig:correct_corrected}
  \end{minipage}
\caption{\textbf{Examples of our motion correction.} Upper:  Legs penetrate each other. Lower: Unstable pose is observed. Synthesized motions (b) are generated from original motions (a) and then rectified (c).}
\label{fig:ik_example}
\vspace{-3mm}
\end{figure}

The IK-based motion editing needs target positions in all frames of motion.
To achieve this semi-automatically, we present an  effortless IK-based motion synthesis that only requires a user to provide a target sampling space $\mathbb{P}$ for the pose of the keyframe $\bm{x}_{\rm key}$ on each action class.
Examples of a kick class are shown in Figs.~\ref{fig:kick_space} and \ref{fig:target_propagation}.
The user determines the target sampling space as shown in Fig.~\ref{fig:kick_space}.
Then, the keyframe for the kick class is defined as the frame where a kicking foot reaches the farthest position from the body.

Given the sampling space for the keyframe, an IK target position $\bm{p}^{\rm sample}_{t_{\rm key}} \in \mathbb{P}$ for the keyframe is randomly sampled.
Target positions $\bm{p}^{\rm target}_t$ for all frames are determined by propagating the difference between $\bm{p}^{\rm sample}_{t_{\rm key}}$ and the end-effector position at the keyframe $\bm{p}_{t_{\rm key}}$ to backward and forward, in a linearly-decreasing manner, as shown in Fig.~\ref{fig:target_propagation} and expressed as follows:
\begin{align}
    \bm{p}^{\rm diff} &= \bm{p}^{\rm sample}_{t_{\rm key}} - \bm{p}_{t_{\rm key}} \nonumber\\
    \bm{p}^{\rm target}_t &= \bm{p}_t + \bm{p}^{\rm diff} \cdot f(t_{\rm key}, t) \nonumber\\
    f(t_{\rm key}, t)    &= \begin{cases}
                \cfrac{t}{t_{\rm key}} &\mbox{ if } t \leq t_{\rm key} \nonumber\\
                \cfrac{T - t}{T - t_{\rm key}} &\mbox{ if } t > t_{\rm key}
    \end{cases}\nonumber\\
    \bm{X}_{\text{aug}} &= \{\text{IK}(\bm{x}_1, \bm{p}^{\rm target}_1)
                                    , \ldots, \text{IK}(\bm{x}_T, \bm{p}^{\rm target}_T)\} \nonumber
\end{align}
where $\wh{\bm{x}} = \text{IK}(\bm{x}, \bm{p})$ is an IK function, and $\bm{x}$ and $\bm{p}$ denote a pose vector and and a 3D position, respectively.
We apply IK with automatically obtained targets $\bm{p}^{\rm target}_{t}$ to all frames for obtaining a synthesized motion $\bm{X}_{\rm aug}$ where the end-effector smoothly reaches $\bm{p}^{\rm sample}_{t_{\rm key}}$.

\subsection{Motion Correction with Imitation Learning using Physics Simulation}
\label{subsec:imitation_learning}

Although most synthesized motions generated by our method are physically realistic, some of them are not.
For example, footskating by VAE, mutual penetrations between body parts by IK, and unstable poses by VAE and IK are empirically observed, as shown in the middle column of Fig.~\ref{fig:ik_example}.

DeepMimic~\cite{peng2018deepmimic} is an imitation learning scheme that allows a physically-simulated character to mimic various motions.
Given a goal motion (\eg, motion measured by a motion capture system), imitation learning trains a policy that modifies a character pose at $t+1$ from its body status at $t$ so that the sequence of the modified poses gets close to the goal motion.
Then, to physically control the character toward the modified pose at each moment, a Proportional-Differential (PD) controller suggests torques given to the character at $t$.
We can obtain the modified motion where the physical character performs by repeating this scheme. 

While, in DeepMimic~\cite{peng2018deepmimic}, the motion is modified for compensating dynamic mismatch (i.e., the difference between the bodies of the goal motion and the character), the goal motion is already physically-plausible because a motion capture system measures it.
On the other hand, we apply this imitation learning to rectify a physically-implausible motion produced by our method.
This physically-implausible motion makes our problem  more challenging because the policy must rectify physical implausibility and the dynamic mismatch.
To cope with this more challenging problem, our imitation learning scheme employs Residual Force Control (RFC)~\cite{yuan2020residual} maintaining physical stability such as fall prevention. With RFC, learnable additional external forces given to the root joint of the character achieve physical stability. 
The rectified motion of the character is still physically-plausible because additional external forces are minimized in training while the pose similarity between a goal motion and the character is maximized.

While DeepMimic using RFC allows us to generate stable motions, the convergence of the training process usually takes more than one day for rectifying one motion with several CPU threads.
This cost is a critical problem when we augment a large number of motions.
The dominant cost in convergence time is on reinforcement learning of the policy network that requires exploration in the policy action space, specifically a target character pose and an additional external force.
%
Although the policy network learns additional external forces from scratch, the learned forces just reduce the positional difference between the character and goal motion.
Based on this observation, we propose the PD-residual force that calculates additional external forces with the PD controller based on the positional difference between the character and goal motion.
This simple modification allows us to omit the learning of external forces and significantly shorten the training process by reducing the dimensionality of the policy action space to explore.

\subsection{Motion Debiasing}
\label{subsec:debiasing}

 \begin{figure}[t]
 \centering
  \begin{minipage}[t]{0.3\linewidth}
    \includegraphics[width=\textwidth]{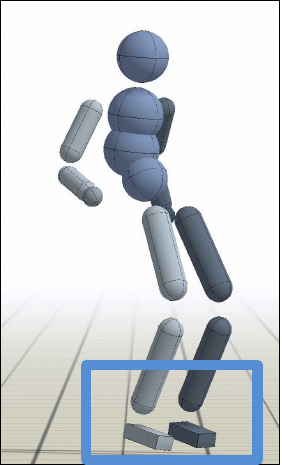}
    \subcaption{original motion}\label{fig/bias_original}
  \end{minipage}
  \begin{minipage}[t]{0.298\linewidth}
    \includegraphics[width=\textwidth]{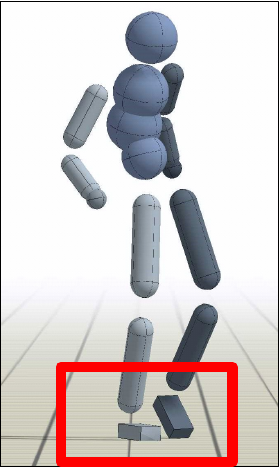}
    \subcaption{rectified motion}\label{fig/bias_phys}
  \end{minipage}
  \begin{minipage}[t]{0.3\linewidth}
    \includegraphics[width=\textwidth]{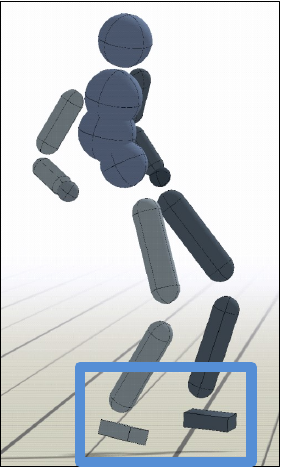}
    \subcaption{debiased motion}\label{fig/bias_debiased}
  \end{minipage}
  \caption{\textbf{Examples of our motion debiasing.} The motion bias is introduced by the dynamic mismatch on the \textit{kick} class motion.}
  \label{fig:motion_debias}
  \vspace{-3mm}
\end{figure}

Our imitation learning scheme explained in Sec.~\ref{subsec:imitation_learning} can rectify synthesized motions to be physically-plausible.
However, a prediction model trained with these rectified motions cannot entirely reduce the prediction error due to the motion bias introduced by the dynamic mismatch~\cite{yuan2020residual} during imitation learning.
The dynamic mismatch is the body difference between ``real humans with hundreds of bones and muscles and deformable skins'' and ``the simulated character with torque-actuated joints and rigid body surfaces.''
Due to this difference, the simulated character fails to fully imitate the motions even if they are physically-plausible, especially motions with fine footwork.

To alleviate the motion bias, we propose motion debiasing to offset the biases imposed by this dynamic mismatch.
We construct training pairs that are original motions and their modified motions by imitation learning as Sec.~\ref{subsec:imitation_learning}. These pairs only contain the bias imposed by the dynamic mismatch.
We propose a simple motion debiasing model of several fully connected layers that map the biased to unbiased motions in framewise.
We apply this motion debiasing to the rectified motions of synthesized motions by our VAE- and IK-based syntheses.
As a result, we obtained the debiased various physically-plausible motions and further improved prediction accuracy, as shown in Fig.~\ref{fig:motion_debias}.

\section{Experiments}

Our experiments consist of four parts: 
Ablation study on our VAE-based method. The effects of several components of our proposed method are validated by the physical and contextual closeness between synthesized and test motions (Sec.~\ref{subsection:vae_motion}).
Convergence time comparison on imitation learning with our PD-residual force (Sec.~\ref{subsection:convergence}).
Performance evaluation on motion prediction with different augmentations (Sec.~\ref{subsection:motion_prediction}).
Augmentation comparison to the previous method DeepMimic~\cite{peng2018deepmimic} (Sec.~\ref{subsection:augumentation}).

\noindent \textbf{Dataset:}
Our experiments were conducted on HDM05 Motion Database~\cite{muller2007documentation}. 
HDM05 is a relatively small and challenging dataset with dynamic motions compared to other standard benchmarks such as Human3.6M~\cite{h36m_pami}. 
We tested our method with five-fold cross-validation where our models were trained on the motion sets of four actors and tested on one of the last actor. 
Motion sets of \textit{punch}, \textit{kick}, and \textit{walk} action classes were resampled to 30Hz and used for the experiments. 
The number of synthesized motions from our VAE and IK is ten times larger than the train set.

\subsection{Motion Synthesis by VAE}
\label{subsection:vae_motion}

The effectiveness of our VAE-based motion synthesis is validated by ablation.
For comparison, a GAN-based method is also evaluated.

\noindent \textbf{Implementation Details:}
All encoders, decoders, and discriminators consist of 256-D LSTM cells and one fully connected layer to output poses. 
The dimension of a latent space is 128 for VAE.
The noise dimension for GAN is also 128.
We used the SGD optimizer to train models for 20,000 epochs.
The number of samples to take mean $n_s=2$ and clusters $n_c=3$ is used for sampling-near-samples. 

\noindent \textbf{Metrics:}
The quality of synthesized motions is evaluated with two metrics: the minimum Dynamic Time Warping (DTW)~\cite{1163055} distance and the Maximum Mean Discrepancy (MMD)~\cite{JMLR:v13:gretton12a}.
The minimum DTW distance is a DTW distance between a test motion and the synthesized motion (training set for original motions) closest to it.
For DTW, frame-wise distances are calculated based on the Euclidean distance in the Euler angle in the radian scale. The sum over all joints except a root joint is evaluated.
MMD measures the distribution distance between the test and synthesized motions.
Minimum DTW distance and MMD measure how the synthesized motions are close to the test motions physically and contextually.
The lower score is better in both metrics.

\begin{table}[t]
\centering
\caption{Quantitative evaluation of augmented motions.}
\begin{tabular}{lccll}
\cline{1-3}
                      & \multicolumn{1}{l}{Min DTW} & \multicolumn{1}{l}{MMD} &  &  \\ \cline{1-3}
Original Motions      & 2.92                              & 0.00                       &  &  \\ \cline{1-3}
GAN                   & 2.90                              & 4.28                    &  &  \\
VAE                   & 2.73                              & 1.94                    &  &  \\
VAE + adv training    & 2.71                              & 1.37                    &  &  \\
VAE + sampl. near samples& 2.72                              & 0.85                    &  &  \\
VAE + both (proposed) & \bf{2.70}                         & \bf{0.20}           &  &  \\ \cline{1-3}
\end{tabular}
\label{tab:motion_generation}
\vspace{-3mm}
\end{table}

\noindent \textbf{Results:}
Table~\ref{tab:motion_generation} shows that the proposed VAE-based method with adversarial training and sampling-near-samples performs best in both metrics. 
Meanwhile, a GAN-based method fails to decrease the minimum DTW distance and gets the highest MMD because the training dataset is too small for GAN to learn various patterns and falls mode collapse problem.

\subsection{Convergence Time Comparison on Imitation Learning}
\label{subsection:convergence}

The convergence times of RFC~\cite{yuan2020residual} and our imitation learning with PD-residual force are evaluated. 

\noindent{\textbf{Implementation details:}}
We use Bullet Physics~\cite{DBLP:conf/siggraph/Coumans15} as the physics engine.
We build the humanoid model from the skeleton of the MDM05 Motion Database, which has 52 DoF and 16 rigid bodies.
We use the same reward function $r_t$ as RFC~\cite{yuan2020residual}.
We train both methods on one kick motion for 100 hours with five threads of Intel® Xeon® Gold 6248 CPU.

\noindent{\textbf{Metrics:}}
We evaluate the training time vs. normalized reward.
The normalized reward is calculated with the obtained reward over the maximum reward on one episode. The normalized reward is calculated based on the motion similarity $r^{\rm im}_t$ and its max value $r_t^{\rm im, max}$.
\begin{align}
    R_{\rm norm} = \cfrac{1}{T}\sum_{t=0}^T \cfrac{r^{\rm im}_t}{r_t^{\rm im, max}} \nonumber
\end{align}

\begin{figure}[t]
\centering
\includegraphics[width=\linewidth]{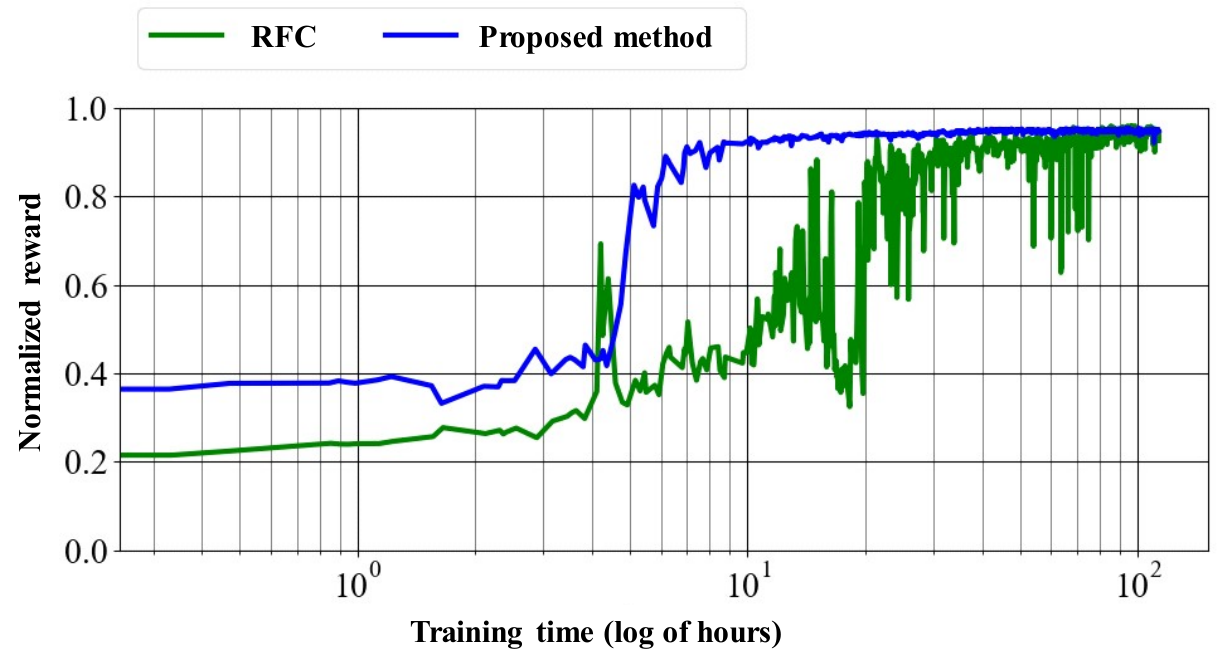}
\caption{\textbf{Normalized rewards vs. training time in logarithmic scale.} }
\label{fig/comp_cvgc}
\vspace{-3mm}
\end{figure}

\noindent{\textbf{Results:}}
The results are shown in Fig.~\ref{fig/comp_cvgc}.
While RFC requires 30 hours for convergence, our method converges around 9 hours
thanks to the dimensionality reduction by our PD-residual force.
Furthermore, our method is stabler than RFC in terms of the reward curve because the physical stability is kept throughout the training process.

\subsection{Motion Prediction with DA}
\label{subsection:motion_prediction}

\noindent \textbf{Implementation details:}
We use the same parameters for VAE as Sec.~\ref{subsection:vae_motion}.
FABRIK~\cite{Aristidou:2011:FABRIK} is used as the IK algorithm.
The keyframes for our IK synthesis are set as the frame when a foot joint reaches the furthest position from the root joint for all action classes.
We chose the foot joint to modify the motion for \textit{punch} class because more motion variation is observed on the foot joint than a hand joint. 
The IK target sampling spaces are set as fan-shapes shown in Fig.~\ref{fig:kick_space} for all action classes.
The parameters for fan-shapes are determined based on the position $\bm{p}_{\text{key}} = (r, h, \theta)$ of the foot joint on the keyframe in cylindrical coordinates.
For \textit{punch}, \textit{kick}, \textit{walk} classes, the IK target positions are sampled from ($[0.5, 2.0]r$, $[1.0, 1.0]h, [-1.7, 1.7]+\theta$), ($[0.8, 1.2]r$, $[0.8, 1.2]h, [-0.785, 0.785]+\theta$), and ($[0.5, 2.0]r$, $[1.0, 1.0]h, [-0.3, 0.3]+\theta$) respectively.
Our motion debiasing network is four 512-dim fully-connected layers with the ReLU activation to offset the bias framewise.
We also temporally expand and shrink motion sequences in the range of 10\% shorter and 10\% longer as temporal data augmentation.

\noindent \textbf{Prediction Model and Metrics:}
We use the heavily benchmarked RNN baseline~\cite{martinez2017human} and the SOTA GCN-based model~\cite{DBLP:conf/iccv/MaoLSL19} to evaluate the effectiveness of our motion DA method on the human motion prediction task. We follow the standard evaluation protocol used in \cite{fragkiadaki2015recurrent,martinez2017human}, 
and report the Euclidean distance between the predicted and ground-truth joint angles in Euler representation. The reported errors in the radian scale are summed over all joints except a root joint and temporally averaged.

\noindent \textbf{Results:}
In Tables~\ref{tab:motion_prediction_GCN} and~\ref{tab:motion_prediction_seq2seq}, we show quantitative results for human motion prediction with data augmentation combinations.
The prediction errors are shown on three timesteps (100, 200, 400ms) for three action classes (\textit{punch}, \textit{kick}, \textit{walk}).
The motion syntheses themselves (rows with no checkmark) often fail to decrease the prediction errors compared to ``No Aug'' because the motion prediction model learns unrealistic motions that are far from test motion data recorded in the real world.
The motion syntheses with physical correction (rows with one checkmark) also fail because the prediction model learns biased motion data different from test motions.
Our proposed motion data augmentation (rows with two checkmarks) achieved the lowest prediction error in all cases by a large margin compared to the previous method ``Noise''.

\subsection{Augmentation Comparison to Previous Method}
\label{subsection:augumentation}

\begin{figure}[t]
\centering
  \begin{minipage}[t]{0.22\linewidth}
    \centering
    \includegraphics[width=\textwidth]{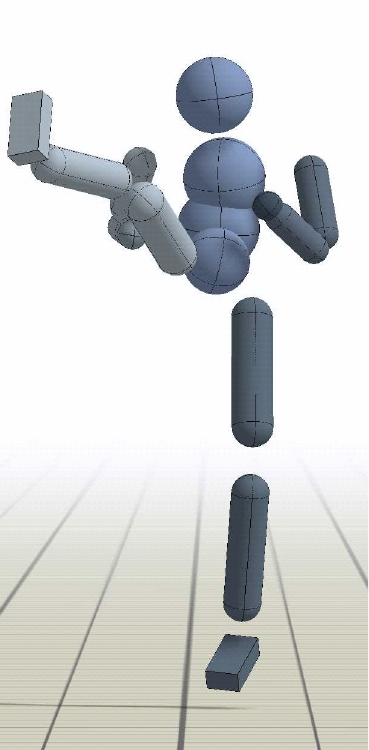}
    \subcaption{original}\label{fig/compDM_original}
  \end{minipage}
  \begin{minipage}[t]{0.22\linewidth}
    \centering
    \includegraphics[width=\textwidth]{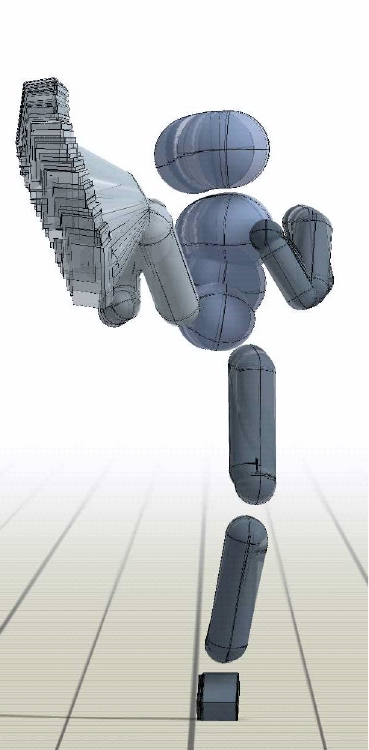}
    \subcaption{DM \#1}\label{fig/compDM_strike}
  \end{minipage}
  \begin{minipage}[t]{0.22\linewidth}
    \centering
    \includegraphics[width=\textwidth]{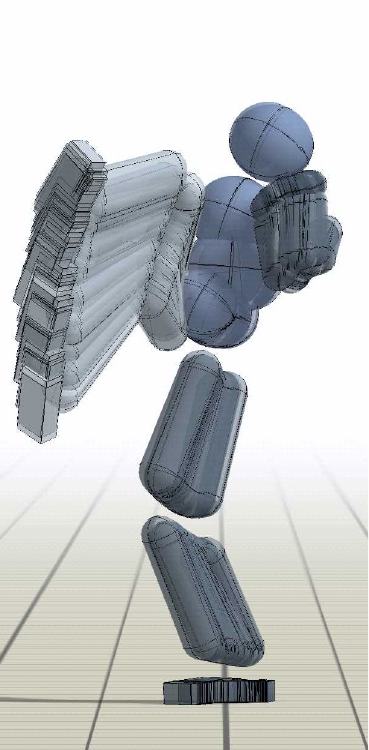}
    \subcaption{DM \#2}\label{fig/compDM_more_strike}
  \end{minipage}
  \begin{minipage}[t]{0.29\linewidth}
    \centering
    \includegraphics[width=\textwidth]{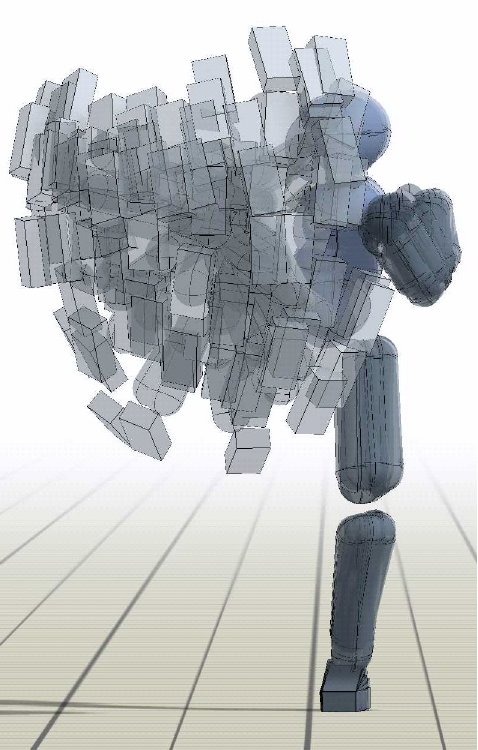}
    \subcaption{proposed}\label{fig/compDM_proposed}
  \end{minipage}
  \caption{\textbf{Augmentation comparison to the base method, DeepMimic\cite{peng2018deepmimic}}. (a) shows an original motion. (b) and (c) show 100 augmented motions by DeepMimic with two rewards weighting $\{\alpha=0.7, \beta=0.3\}$ and $\{\alpha=0.3, \beta=0.7\}$ respectively. (d) shows 100 augmented motions by our method.}\label{fig/compDM}
  \vspace{-3mm}
\end{figure}

\begin{table*}[t]
\centering
\caption{Quantitative results of motion data augmentation on the RNN-based human motion prediction~\cite{martinez2017human}.}
\begin{tabular}{lcc|lllllllll}
\hline
\multicolumn{3}{l|}{Methods} & \multicolumn{9}{l}{Prediction errors$\downarrow$[rad] on each action class \& timesteps [ms]} \\ \hline
{\color[HTML]{333333} } & {\color[HTML]{333333} } & {\color[HTML]{333333} } & \multicolumn{3}{c|}{{\color[HTML]{333333} punch}} & \multicolumn{3}{c|}{{\color[HTML]{333333} kick}} & \multicolumn{3}{c}{{\color[HTML]{333333} walk}} \\
\multirow{-2}{*}{{\color[HTML]{333333} }} & \multirow{-2}{*}{{\color[HTML]{333333} \begin{tabular}[c]{@{}c@{}}physical\\ correction\end{tabular}}} & \multirow{-2}{*}{{\color[HTML]{333333} \begin{tabular}[c]{@{}c@{}}motion\\ debiasing\end{tabular}}} & {\color[HTML]{333333} 100} & {\color[HTML]{333333} 200} & \multicolumn{1}{l|}{{\color[HTML]{333333} 400}} & {\color[HTML]{333333} 100} & {\color[HTML]{333333} 200} & \multicolumn{1}{l|}{{\color[HTML]{333333} 400}} & {\color[HTML]{333333} 100} & {\color[HTML]{333333} 200} & {\color[HTML]{333333} 400} \\ \hline
{\color[HTML]{333333} No aug} & {\color[HTML]{333333} -} & {\color[HTML]{333333} -} & {\color[HTML]{333333} 1.58} & {\color[HTML]{333333} 2.2} & \multicolumn{1}{l|}{{\color[HTML]{333333} 2.59}} & {\color[HTML]{333333} 1.28} & {\color[HTML]{333333} 1.89} & \multicolumn{1}{l|}{{\color[HTML]{333333} 2.45}} & {\color[HTML]{333333} 0.74} & {\color[HTML]{333333} 1.15} & {\color[HTML]{333333} 1.49} \\
{\color[HTML]{333333} Noise} & {\color[HTML]{333333} -} & {\color[HTML]{333333} -} & {\color[HTML]{333333} 1.57} & {\color[HTML]{333333} 2.19} & \multicolumn{1}{l|}{{\color[HTML]{333333} 2.57}} & {\color[HTML]{333333} 1.26} & {\color[HTML]{333333} 1.84} & \multicolumn{1}{l|}{{\color[HTML]{333333} 2.35}} & {\color[HTML]{333333} 0.72} & {\color[HTML]{333333} 1.12} & {\color[HTML]{333333} 1.45} \\ \hline
{\color[HTML]{333333} VAE} & \multicolumn{1}{l}{{\color[HTML]{333333} }} & \multicolumn{1}{l|}{{\color[HTML]{333333} }} & {\color[HTML]{333333} 1.48} & {\color[HTML]{333333} 2.17} & \multicolumn{1}{l|}{{\color[HTML]{333333} 2.71}} & {\color[HTML]{333333} 1.28} & {\color[HTML]{333333} 1.90} & \multicolumn{1}{l|}{{\color[HTML]{333333} 2.46}} & {\color[HTML]{333333} 0.68} & {\color[HTML]{333333} 1.11} & {\color[HTML]{333333} 1.55} \\
{\color[HTML]{333333} IK} & \multicolumn{1}{l}{{\color[HTML]{333333} }} & \multicolumn{1}{l|}{{\color[HTML]{333333} }} & {\color[HTML]{333333} 1.58} & {\color[HTML]{333333} 2.41} & \multicolumn{1}{l|}{{\color[HTML]{333333} 3.14}} & {\color[HTML]{333333} 1.44} & {\color[HTML]{333333} 2.20} & \multicolumn{1}{l|}{{\color[HTML]{333333} 2.95}} & {\color[HTML]{333333} 0.71} & {\color[HTML]{333333} 1.11} & {\color[HTML]{333333} 1.47} \\
{\color[HTML]{333333} VAE \& IK} & \multicolumn{1}{l}{{\color[HTML]{333333} }} & \multicolumn{1}{l|}{{\color[HTML]{333333} }} & {\color[HTML]{333333} 1.56} & {\color[HTML]{333333} 2.38} & \multicolumn{1}{l|}{{\color[HTML]{333333} 3.03}} & {\color[HTML]{333333} 1.21} & {\color[HTML]{333333} 1.78} & \multicolumn{1}{l|}{{\color[HTML]{333333} 2.25}} & {\color[HTML]{333333} 0.67} & {\color[HTML]{333333} 1.07} & {\color[HTML]{333333} 1.43} \\ \hline
{\color[HTML]{333333} VAE} & {\color[HTML]{333333} \checkmark} & {\color[HTML]{333333} } & {\color[HTML]{333333} 1.57} & {\color[HTML]{333333} 2.19} & \multicolumn{1}{l|}{{\color[HTML]{333333} 2.75}} & {\color[HTML]{333333} 1.27} & {\color[HTML]{333333} 1.92} & \multicolumn{1}{l|}{{\color[HTML]{333333} 2.57}} & {\color[HTML]{333333} 0.71} & {\color[HTML]{333333} 1.14} & {\color[HTML]{333333} 1.56} \\
{\color[HTML]{333333} IK} & {\color[HTML]{333333} \checkmark} & {\color[HTML]{333333} } & {\color[HTML]{333333} 1.52} & {\color[HTML]{333333} 2.26} & \multicolumn{1}{l|}{{\color[HTML]{333333} 3.02}} & {\color[HTML]{333333} 1.26} & {\color[HTML]{333333} 1.94} & \multicolumn{1}{l|}{{\color[HTML]{333333} 2.59}} & {\color[HTML]{333333} 0.74} & {\color[HTML]{333333} 1.22} & {\color[HTML]{333333} 1.78} \\
{\color[HTML]{333333} VAE \& IK} & {\color[HTML]{333333} \checkmark} & {\color[HTML]{333333} } & {\color[HTML]{333333} 1.53} & {\color[HTML]{333333} 2.20} & \multicolumn{1}{l|}{{\color[HTML]{333333} 2.97}} & {\color[HTML]{333333} 1.24} & {\color[HTML]{333333} 1.91} & \multicolumn{1}{l|}{{\color[HTML]{333333} 2.60}} & {\color[HTML]{333333} 0.71} & {\color[HTML]{333333} 1.19} & {\color[HTML]{333333} 1.71} \\ \hline
{\color[HTML]{333333} VAE} & {\color[HTML]{333333} \checkmark} & {\color[HTML]{333333} \checkmark} & {\color[HTML]{333333} 1.50} & {\color[HTML]{333333} 2.10} & \multicolumn{1}{l|}{{\color[HTML]{333333} 2.63}} & {\color[HTML]{333333} 1.11} & {\color[HTML]{333333} 1.64} & \multicolumn{1}{l|}{{\color[HTML]{333333} 2.06}} & {\color[HTML]{333333} 0.66} & {\color[HTML]{333333} 1.08} & {\color[HTML]{333333} 1.48} \\
{\color[HTML]{333333} IK} & {\color[HTML]{333333} \checkmark} & {\color[HTML]{333333} \checkmark} & {\color[HTML]{333333} 1.39} & {\color[HTML]{333333} 1.93} & \multicolumn{1}{l|}{{\color[HTML]{333333} \textbf{2.45}}} & {\color[HTML]{333333} 1.05} & {\color[HTML]{333333} 1.52} & \multicolumn{1}{l|}{{\color[HTML]{333333} 1.85}} & {\color[HTML]{333333} \textbf{0.57}} & {\color[HTML]{333333} \textbf{0.91}} & {\color[HTML]{333333} \textbf{1.22}} \\
{\color[HTML]{333333} VAE \& IK} & {\color[HTML]{333333} \checkmark} & {\color[HTML]{333333} \checkmark} & {\color[HTML]{333333} \textbf{1.37}} & {\color[HTML]{333333} \textbf{1.91}} & \multicolumn{1}{l|}{{\color[HTML]{333333} 2.49}} & {\color[HTML]{333333} \textbf{1.01}} & {\color[HTML]{333333} \textbf{1.48}} & \multicolumn{1}{l|}{{\color[HTML]{333333} \textbf{1.80}}} & {\color[HTML]{333333} 0.58} & {\color[HTML]{333333} 0.92} & {\color[HTML]{333333} 1.24} \\ \hline
\end{tabular}
\label{tab:motion_prediction_GCN}
\vspace{-3mm}
\end{table*}

\begin{table*}[t]
\centering
\caption{Quantitative results of motion data augmentation on the SOTA GCN-based human motion prediction~\cite{DBLP:conf/iccv/MaoLSL19}.}
\begin{tabular}{lcc|lllllllll}
\hline
\multicolumn{3}{l|}{Methods} & \multicolumn{9}{l}{Prediction errors$\downarrow$[rad] on each action class \& timesteps [ms]} \\ \hline
{\color[HTML]{333333} } & {\color[HTML]{333333} } & {\color[HTML]{333333} } & \multicolumn{3}{c|}{{\color[HTML]{333333} punch}} & \multicolumn{3}{c|}{{\color[HTML]{333333} kick}} & \multicolumn{3}{c}{{\color[HTML]{333333} walk}} \\
\multirow{-2}{*}{{\color[HTML]{333333} }} & \multirow{-2}{*}{{\color[HTML]{333333} \begin{tabular}[c]{@{}c@{}}physical\\ correction\end{tabular}}} & \multirow{-2}{*}{{\color[HTML]{333333} \begin{tabular}[c]{@{}c@{}}motion\\ debiasing\end{tabular}}} & {\color[HTML]{333333} 100} & {\color[HTML]{333333} 200} & \multicolumn{1}{l|}{{\color[HTML]{333333} 400}} & {\color[HTML]{333333} 100} & {\color[HTML]{333333} 200} & \multicolumn{1}{l|}{{\color[HTML]{333333} 400}} & {\color[HTML]{333333} 100} & {\color[HTML]{333333} 200} & {\color[HTML]{333333} 400} \\ \hline
{\color[HTML]{333333} No aug} & {\color[HTML]{333333} -} & {\color[HTML]{333333} -} & {\color[HTML]{333333} 1.31} & {\color[HTML]{333333} 1.87} & \multicolumn{1}{l|}{{\color[HTML]{333333} 2.33}} & {\color[HTML]{333333} 1.08} & {\color[HTML]{333333} 1.68} & \multicolumn{1}{l|}{{\color[HTML]{333333} 2.26}} & {\color[HTML]{333333} 0.52} & {\color[HTML]{333333} 0.88} & {\color[HTML]{333333} 1.24} \\
{\color[HTML]{333333} Noise} & {\color[HTML]{333333} -} & {\color[HTML]{333333} -} & {\color[HTML]{333333} 1.31} & {\color[HTML]{333333} 1.90} & \multicolumn{1}{l|}{{\color[HTML]{333333} 2.35}} & {\color[HTML]{333333} 1.06} & {\color[HTML]{333333} 1.65} & \multicolumn{1}{l|}{{\color[HTML]{333333} 2.25}} & {\color[HTML]{333333} 0.52} & {\color[HTML]{333333} 0.87} & {\color[HTML]{333333} 1.21} \\ \hline
{\color[HTML]{333333} VAE} & \multicolumn{1}{l}{{\color[HTML]{333333} }} & \multicolumn{1}{l|}{{\color[HTML]{333333} }} & {\color[HTML]{333333} 1.28} & {\color[HTML]{333333} 1.88} & \multicolumn{1}{l|}{{\color[HTML]{333333} 2.34}} & {\color[HTML]{333333} 1.06} & {\color[HTML]{333333} 1.63} & \multicolumn{1}{l|}{{\color[HTML]{333333} 2.17}} & {\color[HTML]{333333} 0.52} & {\color[HTML]{333333} 0.91} & {\color[HTML]{333333} 1.28} \\
{\color[HTML]{333333} IK} & \multicolumn{1}{l}{{\color[HTML]{333333} }} & \multicolumn{1}{l|}{{\color[HTML]{333333} }} & {\color[HTML]{333333} 1.21} & {\color[HTML]{333333} 1.73} & \multicolumn{1}{l|}{{\color[HTML]{333333} 2.25}} & {\color[HTML]{333333} 0.96} & {\color[HTML]{333333} 1.39} & \multicolumn{1}{l|}{{\color[HTML]{333333} 1.73}} & {\color[HTML]{333333} 0.50} & {\color[HTML]{333333} 0.85} & {\color[HTML]{333333} 1.18} \\
{\color[HTML]{333333} VAE \& IK} & \multicolumn{1}{l}{{\color[HTML]{333333} }} & \multicolumn{1}{l|}{{\color[HTML]{333333} }} & {\color[HTML]{333333} 1.22} & {\color[HTML]{333333} 1.81} & \multicolumn{1}{l|}{{\color[HTML]{333333} 2.29}} & {\color[HTML]{333333} 0.95} & {\color[HTML]{333333} 1.38} & \multicolumn{1}{l|}{{\color[HTML]{333333} 1.71}} & {\color[HTML]{333333} 0.49} & {\color[HTML]{333333} 0.85} & {\color[HTML]{333333} 1.20} \\ \hline
{\color[HTML]{333333} VAE} & {\color[HTML]{333333} \checkmark} & {\color[HTML]{333333} } & {\color[HTML]{333333} 1.31} & {\color[HTML]{333333} 1,89} & \multicolumn{1}{l|}{{\color[HTML]{333333} 2.36}} & {\color[HTML]{333333} 1.03} & {\color[HTML]{333333} 1.60} & \multicolumn{1}{l|}{{\color[HTML]{333333} 2.14}} & {\color[HTML]{333333} 0.52} & {\color[HTML]{333333} 0.89} & {\color[HTML]{333333} 1.25} \\
{\color[HTML]{333333} IK} & {\color[HTML]{333333} \checkmark} & {\color[HTML]{333333} } & {\color[HTML]{333333} 1.31} & {\color[HTML]{333333} 1.89} & \multicolumn{1}{l|}{{\color[HTML]{333333} 2.49}} & {\color[HTML]{333333} 1.06} & {\color[HTML]{333333} 1.66} & \multicolumn{1}{l|}{{\color[HTML]{333333} 2.20}} & {\color[HTML]{333333} 0.53} & {\color[HTML]{333333} 0.94} & {\color[HTML]{333333} 1.35} \\
{\color[HTML]{333333} VAE \& IK} & {\color[HTML]{333333} \checkmark} & {\color[HTML]{333333} } & {\color[HTML]{333333} 1.28} & {\color[HTML]{333333} 1.84} & \multicolumn{1}{l|}{{\color[HTML]{333333} 2.35}} & {\color[HTML]{333333} 1.03} & {\color[HTML]{333333} 1.65} & \multicolumn{1}{l|}{{\color[HTML]{333333} 2.17}} & {\color[HTML]{333333} 0.54} & {\color[HTML]{333333} 0.94} & {\color[HTML]{333333} 1.37} \\ \hline
{\color[HTML]{333333} VAE} & {\color[HTML]{333333} \checkmark} & {\color[HTML]{333333} \checkmark} & {\color[HTML]{333333} \textbf{1.22}} & {\color[HTML]{333333} 1.74} & \multicolumn{1}{l|}{{\color[HTML]{333333} 2.07}} & {\color[HTML]{333333} 1.00} & {\color[HTML]{333333} 1.52} & \multicolumn{1}{l|}{{\color[HTML]{333333} 1.89}} & {\color[HTML]{333333} 0.52} & {\color[HTML]{333333} 0.89} & {\color[HTML]{333333} 1.25} \\
{\color[HTML]{333333} IK} & {\color[HTML]{333333} \checkmark} & {\color[HTML]{333333} \checkmark} & {\color[HTML]{333333} 1.27} & {\color[HTML]{333333} 1.79} & \multicolumn{1}{l|}{{\color[HTML]{333333} 2.24}} & {\color[HTML]{333333} 0.92} & {\color[HTML]{333333} 1.35} & \multicolumn{1}{l|}{{\color[HTML]{333333} 1.65}} & {\color[HTML]{333333} 0.48} & {\color[HTML]{333333} 0.81} & {\color[HTML]{333333} \textbf{1.11}} \\
{\color[HTML]{333333} VAE \& IK} & {\color[HTML]{333333} \checkmark} & {\color[HTML]{333333} \checkmark} & {\color[HTML]{333333} \textbf{1.22}} & {\color[HTML]{333333} \textbf{1.70}} & \multicolumn{1}{l|}{{\color[HTML]{333333} \textbf{2.06}}} & {\color[HTML]{333333} \textbf{0.90}} & {\color[HTML]{333333} \textbf{1.32}} & \multicolumn{1}{l|}{{\color[HTML]{333333} \textbf{1.60}}} & {\color[HTML]{333333} \textbf{0.47}} & {\color[HTML]{333333} \textbf{0.80}} & {\color[HTML]{333333} \textbf{1.11}} \\ \hline
\end{tabular}
\label{tab:motion_prediction_seq2seq}
\vspace{-3mm}
\end{table*}

We compared the augmentation capability of our method and the additional tasks of DeepMimic~\cite{peng2018deepmimic}. 

\noindent{\textbf{Experimental set up:}}
We choose one \textit{kick} class motion and independently augment it with DeepMimic, and our IK-based motion synthesis with motion correction.
DeepMimic can also augment motions by training characters to solve additional tasks besides the original imitation.
The used additional task is defined as DeepMimic's \textit{Strike} reward $r^{\text{strike}}_{t}$ that rewards the character when the foot strikes randomly placed targets.
The targets are randomly placed within the same target sampling space used in our IK-based motion synthesis.
DeepMimic is tested in two rewards weighting $\{\alpha=0.7, \beta=0.3\}$ and $\{\alpha=0.3, \beta=0.7\}$ in the following equation:
\begin{align}
r = \alpha r_{t} + \beta r^{\text{strike}}_{t}
\end{align}

\noindent{\textbf{Results:}}
We show the results in Fig.~\ref{fig/compDM}.
Augmented motions from DeepMimic have limited diversity in both reward weightings because the policy suffers from the tradeoff between the imitation and the additional strike tasks.
Although more weight on the strike reward slightly improves the diversity, the resulting motions lose the original motion details.
On the other hand, our method produces diverse motions by dividing the augmentation to the synthesis and physical correction where the policy focuses only on the imitation task.

\section{Limitations}
Our motion augmentation has two limitations.

First, motion correction still takes several hours to rectify one motion, even with our proposal to accelerate the training. 
This cost makes it hard to apply our motion augmentation to more extensive motion prediction benchmarks such as Human3.6M~\cite{h36m_pami} due to computational cost.
However, the training time could be shortened by using meta-learning~\cite{finn2017model} for better policy initialization or the fast physics simulation environment accelerated with GPU rather than CPU.


Second, our motion augmentation is not immediately applicable to the partially-observed motion sequences, such as only observed upper body or 2D motion sequences, because our motion correction only accepts 3D motion sequences for a whole-body 3D character.
We need to estimate missing joints or 2D-3D pose lifting~\cite{tome2017lifting} to apply our method to these situations.

\vspace{-2mm}
\section{Conclusion}

This work presented a new human motion augmentation approach using VAE- and IK-based motion syntheses and motion correction with physics simulation.
Experiments demonstrated that our augmentation outperformed previous methods because our VAE- and IK-based motion syntheses improve the diversity of training motion data, and our motion correction rectifies the unrealistic artifacts without motion biases.
Our future work includes a new motion synthesis approach and faster motion correction based on meta-learning and GPU acceleration for larger-scale datasets.

{\small
\bibliographystyle{ieee_fullname}
\bibliography{ReviewTemplate}
}

\clearpage

\section{Dataset}
In the following, we give detailed information about the HDM05 Motion Database~\cite{muller2007documentation} and post-processing for our experiments.
HDM05 contains dynamic motion sequences such as \textit{kick}, \textit{punch}, and \textit{jump} classes for 50 minutes in total.
On the other hand, the standard benchmark Human3.6m~\cite{h36m_pami} contains relatively static motions such as \textit{eating}, \textit{talking on the phone}, and \textit{smoking} classes for about 1200 minutes in total.
We find HDM05 more challenging due to dynamic motions and fewer data.
Therefore, HDM05 is suitable for validating our motion data augmentation.
We followed the post-processing procedure of a motion synthesis method~\cite{holden2016deep} that cuts long motion sequences to clips of each motion class and retargets them to the uniform skeleton based on CMU Mocap~\cite{de2009guide} for VAE to learn motions independently from skeletal differences.

\section{Additional Experiments on PD-residual Forces}
In Sec. 4.2, the convergence time comparison is shown only on the \textit{kick} class motion. We further conducted the convergence time comparison on the \textit{walk} and \textit{punch} class motions used to compare motion data augmentation Sec. 4.3.

\noindent \textbf{Implementation Details:}
We use the same setting as Sec. 4.2 to train imitation learning.

\noindent \textbf{Results:}
\begin{figure}
    \centering
    \begin{minipage}[t]{\linewidth}
        \centering
        \includegraphics[width=0.95\linewidth]{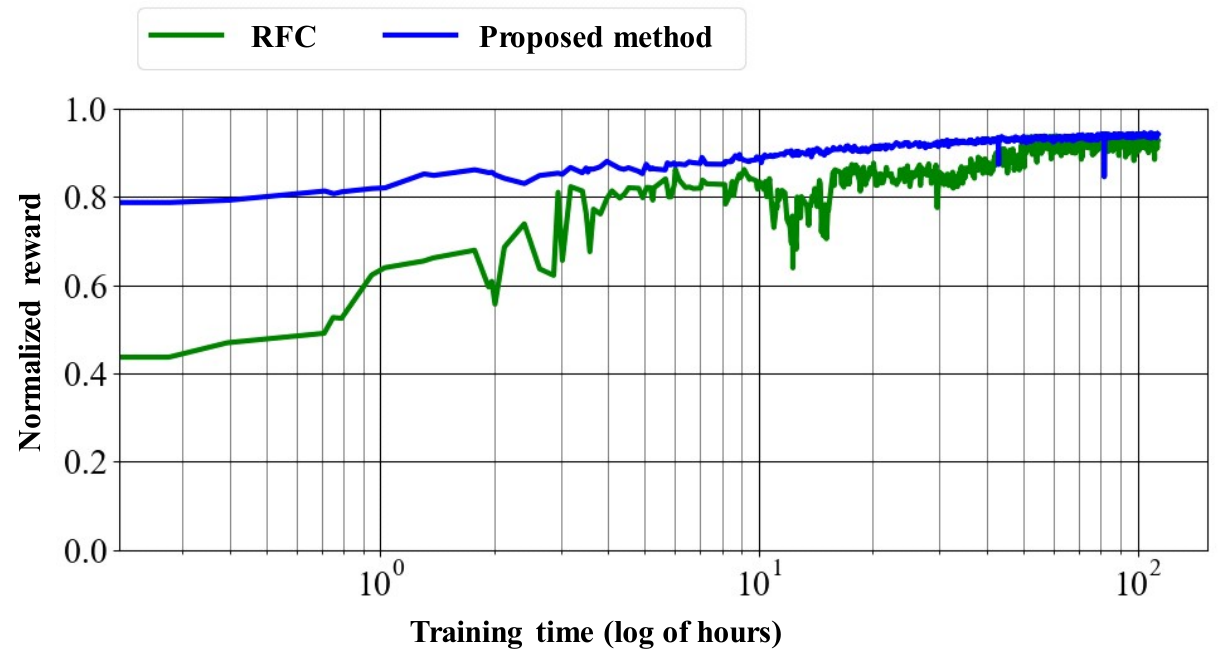}
        \subcaption{comparison on the walk motion.}
    \end{minipage} \\
    \begin{minipage}[t]{\linewidth}
        \centering
        \includegraphics[width=0.95\linewidth]{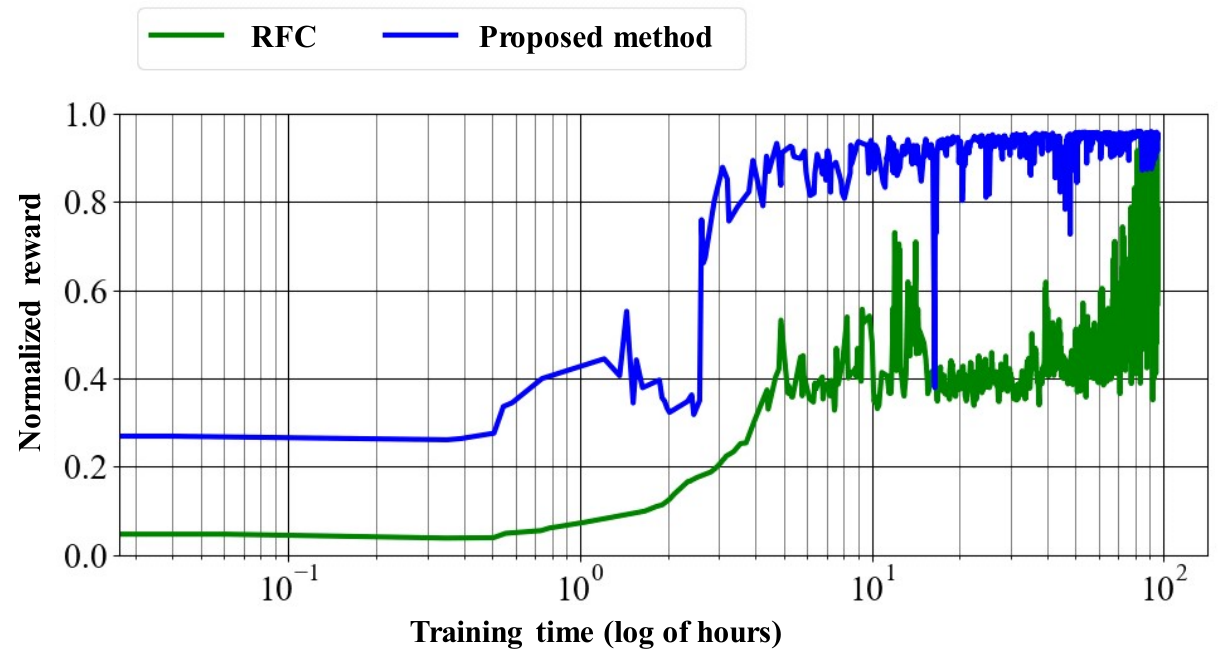}
        \subcaption{comparison on the punch motion.}
    \end{minipage}
    \caption{\textbf{Normalized rewards vs. training time in logarithmic scale.}}
    \label{fig:exp_conv_additional}
\end{figure}
The results are shown in Fig.~\ref{fig:exp_conv_additional}.
Again, imitation learning with our PD-residual forces converges faster and stabler than RFC~\cite{yuan2020residual}.

\section{Visualization of Sampling-near-samples}

\begin{figure*}
    \centering
    \includegraphics[width=0.95\textwidth]{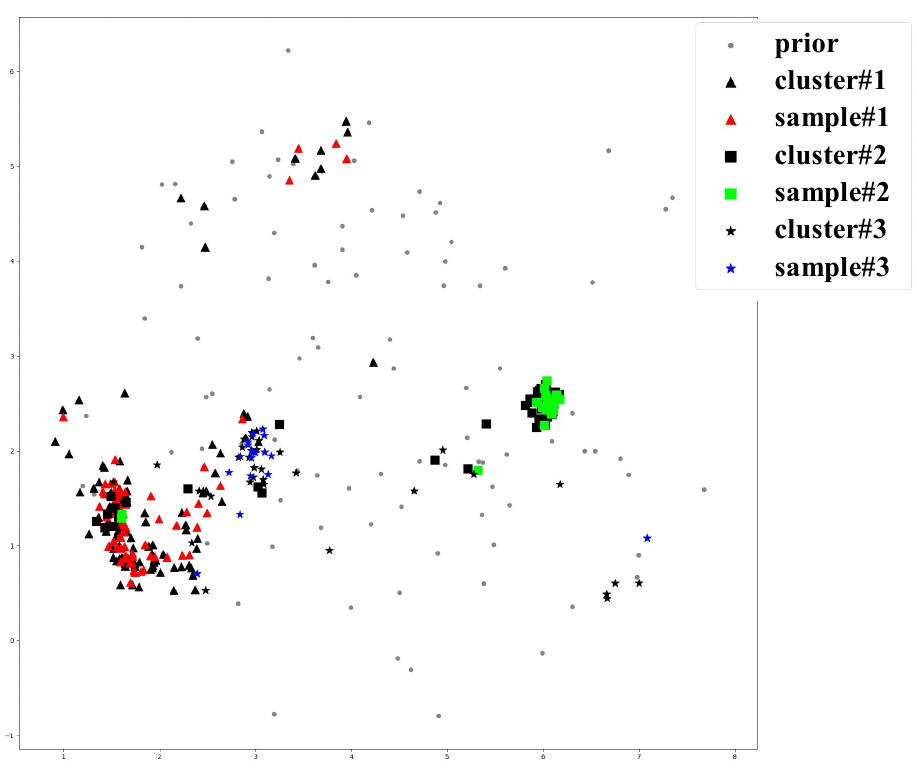}
    \caption{\textbf{Visualization of the sampled latent variables from the prior distribution and proposed sampling-near-samples.} We cluster the train set to three clusters denoted by black triangles($\blacktriangle$), squares($\blacksquare$), and stars($\star$).The prior distribution $\mathcal{N}(\bm{0}, \bm{I})$ samples the latent variables denoted by gray dots($\color{gray} \bullet$) distributed on the unlearned regions that the train set does not cover. The samples from each cluster($\color{red} \blacktriangle, \color{green} \blacksquare, \color{blue} \star$) by our sampling-near-samples locate near the train set and succeed to synthesize dynamic motions even with inefficient data.}
    \label{fig:sns_distribution}
\end{figure*}

We visualized the sampled latent variables from the prior distribution $\mathcal{N}(\bm{0}, \bm{I})$ and proposed sampling-near-samples.

\noindent \textbf{Implementation Details:}
We applied dimension reduction by PCA and subsequent UMAP~\cite{DBLP:journals/corr/abs-1802-03426} to map the latent variables to $\text{dim}=13$ and $\text{dim}=2$ respectively.

\noindent \textbf{Results:}
The plot of the latent variables is shown in Fig.~\ref{fig:sns_distribution}.
One can observe that our sampling-near-samples method successfully samples representations($\color{red} \blacktriangle, \color{green} \blacksquare, \color{blue} \star$) located near the train set ($\blacktriangle$, $\blacksquare$, $\star$).
However, the prior distribution samples representations ($\color{gray} \bullet$) from unlearned regions that the train set does not cover.
As the visualization suggests, our sampling-near-samples method is robust to the sparsity by sampling from the reliably learned regions.

\section{Experiments on Additional Action Classes}
\label{sec:additional_actions}
We conducted further experiments on augmentation for human motion prediction to verify the effectiveness of our approach. 
Five action classes (\textit{grab}, \textit{deposit}, \textit{jog}, \textit{sneak}, \textit{throw}) are added for evaluation.
We chose the hand joint for \textit{grab}, \textit{deposit}, and \textit{throw} to modify the motion sequences.
For \textit{grab}, \textit{deposit}, \textit{jog}, \textit{sneak}, \textit{throw} classes, the IK target positions are sampled from ($[0.5, 2.0]r$, $[1.0, 1.0]h, [-1.7, 1.7]+\theta$), ($[0.5, 2.0]r$, $[1.0, 1.0]h, [-1.7, 1.7]+\theta$), ($[0.5, 2.0]r$, $[1.0, 1.0]h, [-0.3, 0.3]+\theta$), ($[0.5, 2.0]r$, $[1.0, 1.0]h, [-0.3, 0.3]+\theta$), and ($[0.5, 2.0]r$, $[1.0, 1.0]h, [-1.7, 1.7]+\theta$) respectively.
Other experimental settings except action classes follow Sec 4.3.
The results are shown in Tables~\ref{tab:seq2seq_gdjst} and ~\ref{tab:gcn_gdjst}.
Our proposed method outperforms in most cases on the RNN-based model.
However, on the GCN-based model, our motion generation achieves the best performance in most cases.
We can still optimize the augmentation parameters for better performances of the GCN-based model.

\begin{table*}[t]
\centering
\caption{Quantitative results of motion data augmentation on the RNN-based human motion prediction~\cite{martinez2017human} on \textit{grab}, \textit{deposit}, \textit{jog}, \textit{sneak}, \textit{throw}.}
\begin{tabular}{lcc|llllllllll}
\hline
\multicolumn{3}{l|}{Methods} & \multicolumn{10}{c}{action class \& timesteps [ms]} \\ \hline
\multicolumn{1}{c}{\multirow{2}{*}{}} & \multirow{2}{*}{\begin{tabular}[c]{@{}c@{}}physical\\ correction\end{tabular}} & \multirow{2}{*}{\begin{tabular}[c]{@{}c@{}}motion\\ debiasing\end{tabular}} & \multicolumn{2}{c|}{grab$\downarrow$} & \multicolumn{2}{c|}{deposit$\downarrow$} & \multicolumn{2}{c|}{jog$\downarrow$} & \multicolumn{2}{c|}{sneak$\downarrow$} & \multicolumn{2}{c}{throw$\downarrow$} \\
\multicolumn{1}{c}{} &  &  & 100 & \multicolumn{1}{l|}{400} & 100 & \multicolumn{1}{l|}{400} & 100 & \multicolumn{1}{l|}{400} & \multicolumn{1}{c}{100} & \multicolumn{1}{l|}{400} & \multicolumn{1}{c}{100} & 400 \\ \hline
No aug & - & - & 0.71 & \multicolumn{1}{l|}{1.92} & 0.72 & \multicolumn{1}{l|}{\textbf{1.88}} & 0.91 & \multicolumn{1}{l|}{1.67} & 0.46 & \multicolumn{1}{l|}{1.27} & 1.24 & \textbf{2.61} \\
Noise & - & - & 0.71 & \multicolumn{1}{l|}{1.92} & 0.69 & \multicolumn{1}{l|}{1.92} & 0.91 & \multicolumn{1}{l|}{1.65} & 0.45 & \multicolumn{1}{l|}{1.21} & 1.24 & 2.56 \\ \hline
VAE & \multicolumn{1}{l}{} & \multicolumn{1}{l|}{} & 0.83 & \multicolumn{1}{l|}{2.06} & 0.83 & \multicolumn{1}{l|}{2.18} & 0.97 & \multicolumn{1}{l|}{1.79} & 0.53 & \multicolumn{1}{l|}{1.50} & \textbf{1.20} & \textbf{2.61} \\
IK & \multicolumn{1}{l}{} & \multicolumn{1}{l|}{} & 0.89 & \multicolumn{1}{l|}{2.26} & 0.85 & \multicolumn{1}{l|}{2.22} & 0.94 & \multicolumn{1}{l|}{1.82} & 0.55 & \multicolumn{1}{l|}{1.46} & 1.46 & 3.16 \\
VAE \& IK & \multicolumn{1}{l}{} & \multicolumn{1}{l|}{} & 0.78 & \multicolumn{1}{l|}{1.99} & 0.78 & \multicolumn{1}{l|}{2.08} & 0.90 & \multicolumn{1}{l|}{1.80} & 0.46 & \multicolumn{1}{l|}{1.38} & 1.27 & 2.81 \\ \hline
VAE & \checkmark & \multicolumn{1}{l|}{} & 0.84 & \multicolumn{1}{l|}{2.25} & 0.72 & \multicolumn{1}{l|}{2.04} & 1.15 & \multicolumn{1}{l|}{2.27} & 0.55 & \multicolumn{1}{l|}{1.57} & 1.25 & 2.71 \\
IK & \checkmark & \multicolumn{1}{l|}{} & 0.88 & \multicolumn{1}{l|}{2.23} & 0.86 & \multicolumn{1}{l|}{2.23} & 0.96 & \multicolumn{1}{l|}{2.14} & 0.54 & \multicolumn{1}{l|}{1.56} & 1.41 & 3.16 \\
VAE \& IK & \checkmark &  & 0.87 & \multicolumn{1}{l|}{2.30} & 0.85 & \multicolumn{1}{l|}{2.31} & 0.99 & \multicolumn{1}{l|}{2.35} & 0.49 & \multicolumn{1}{l|}{1.50} & 1.40 & 3.19 \\ \hline
VAE & \checkmark & \checkmark & 0.69 & \multicolumn{1}{l|}{1.87} & 0.71 & \multicolumn{1}{l|}{1.92} & 0.87 & \multicolumn{1}{l|}{1.71} & 0.43 & \multicolumn{1}{l|}{1.26} & 1.21 & \textbf{2.61} \\
IK & \checkmark & \checkmark & 0.70 & \multicolumn{1}{l|}{1.81} & 0.74 & \multicolumn{1}{l|}{1.99} & \textbf{0.73} & \multicolumn{1}{l|}{\textbf{1.40}} & 0.41 & \multicolumn{1}{l|}{1.16} & 1.37 & 2.93 \\
VAE \& IK & \checkmark & \checkmark & \textbf{0.64} & \multicolumn{1}{l|}{\textbf{1.68}} & \textbf{0.72} & \multicolumn{1}{l|}{1.94} & 0.77 & \multicolumn{1}{l|}{1.50} & \textbf{0.39} & \multicolumn{1}{l|}{\textbf{1.14}} & 1.31 & 2.87 \\ \hline
\end{tabular}
\label{tab:seq2seq_gdjst}
\end{table*}

\begin{table*}[t]
\centering
\caption{Quantitative results of motion data augmentation on the SOTA GCN-based human motion prediction~\cite{DBLP:conf/iccv/MaoLSL19} on \textit{grab}, \textit{deposit}, \textit{jog}, \textit{sneak}, \textit{throw}.}
\begin{tabular}{lcc|llllllllll}
\hline
\multicolumn{3}{l|}{Methods} & \multicolumn{10}{c}{action class \& timesteps [ms]} \\ \hline
\multicolumn{1}{c}{\multirow{2}{*}{}} & \multirow{2}{*}{\begin{tabular}[c]{@{}c@{}}physical\\ correction\end{tabular}} & \multirow{2}{*}{\begin{tabular}[c]{@{}c@{}}motion\\ debiasing\end{tabular}} & \multicolumn{2}{c|}{grab$\downarrow$} & \multicolumn{2}{c|}{deposit$\downarrow$} & \multicolumn{2}{c|}{jog$\downarrow$} & \multicolumn{2}{c|}{sneak$\downarrow$} & \multicolumn{2}{c}{throw$\downarrow$} \\
\multicolumn{1}{c}{} &  &  & 100 & \multicolumn{1}{l|}{400} & 100 & \multicolumn{1}{l|}{400} & 100 & \multicolumn{1}{l|}{400} & \multicolumn{1}{c}{100} & \multicolumn{1}{l|}{400} & \multicolumn{1}{c}{100} & 400 \\ \hline
No aug & - & - & 0.56 & \multicolumn{1}{l|}{1.79} & 0.55 & \multicolumn{1}{l|}{\textbf{1.71}} & 0.66 & \multicolumn{1}{l|}{1.33} & 0.27 & \multicolumn{1}{l|}{0.94} & 1.03 & 2.45 \\
Noise & - & - & 0.56 & \multicolumn{1}{l|}{1.79} & 0.55 & \multicolumn{1}{l|}{1.70} & 0.66 & \multicolumn{1}{l|}{1.31} & 0.27 & \multicolumn{1}{l|}{0.93} & 1.03 & 2.40 \\ \hline
VAE & \multicolumn{1}{l}{} & \multicolumn{1}{l|}{} & 0.52 & \multicolumn{1}{l|}{1.65} & 0.53 & \multicolumn{1}{l|}{\textbf{1.68}} & 0.67 & \multicolumn{1}{l|}{1.32} & 0.30 & \multicolumn{1}{l|}{0.95} & 1.06 & \textbf{2.28} \\
IK & \multicolumn{1}{l}{} & \multicolumn{1}{l|}{} & 0.58 & \multicolumn{1}{l|}{1.79} & 0.56 & \multicolumn{1}{l|}{1.76} & 0.62 & \multicolumn{1}{l|}{\textbf{1.24}} & 0.29 & \multicolumn{1}{l|}{0.92} & 1.08 & 2.45 \\
VAE \& IK & \multicolumn{1}{l}{} & \multicolumn{1}{l|}{} & 0.53 & \multicolumn{1}{l|}{1.69} & \textbf{0.52} & \multicolumn{1}{l|}{\textbf{1.68}} & \textbf{0.61} & \multicolumn{1}{l|}{1.26} & 0.27 & \multicolumn{1}{l|}{\textbf{0.90}} & \textbf{1.02} & 2.31 \\ \hline
VAE & \checkmark & \multicolumn{1}{l|}{} & 0.59 & \multicolumn{1}{l|}{1.80} & 0.57 & \multicolumn{1}{l|}{1.74} & 0.72 & \multicolumn{1}{l|}{1.49} & 0.33 & \multicolumn{1}{l|}{1.06} & 1.05 & 2.43 \\
IK & \checkmark & \multicolumn{1}{l|}{} & 0.55 & \multicolumn{1}{l|}{1.77} & 0.53 & \multicolumn{1}{l|}{1.76} & 0.69 & \multicolumn{1}{l|}{1.51} & 0.30 & \multicolumn{1}{l|}{1.09} & 1.07 & 2.57 \\
VAE \& IK & \checkmark &  & 0.55 & \multicolumn{1}{l|}{1.80} & 0.54 & \multicolumn{1}{l|}{1.74} & 0.73 & \multicolumn{1}{l|}{1.68} & 0.30 & \multicolumn{1}{l|}{1.10} & 1.05 & 2.47 \\ \hline
VAE & \checkmark & \checkmark & 0.51 & \multicolumn{1}{l|}{1.64} & 0.55 & \multicolumn{1}{l|}{1.77} & 0.72 & \multicolumn{1}{l|}{1.45} & 0.29 & \multicolumn{1}{l|}{1.03} & 1.09 & 2.37 \\
IK & \checkmark & \checkmark & 0.52 & \multicolumn{1}{l|}{1.67} & 0.55 & \multicolumn{1}{l|}{1.73} & 0.63 & \multicolumn{1}{l|}{1.28} & 0.28 & \multicolumn{1}{l|}{0.98} & 1.08 & 2.39 \\
VAE \& IK & \checkmark & \checkmark & \textbf{0.48} & \multicolumn{1}{l|}{\textbf{1.60}} & 0.53 & \multicolumn{1}{l|}{1.70} & 0.64 & \multicolumn{1}{l|}{1.34} & \textbf{0.26} & \multicolumn{1}{l|}{\textbf{0.90}} & 1.06 & 2.39 \\ \hline
\end{tabular}
\label{tab:gcn_gdjst}
\end{table*}

\section{Experiments on Transformer-based Human Motion Prediction Model}
We further validated the effectiveness of our approach on the transformer-based human motion prediction model~\cite{DBLP:conf/iccvw/Martinez-Gonzalez21}.
Other experimental settings except the prediction model follow Sec 4.3 and Sec~\ref{sec:additional_actions}.
The results are shown in Tables~\ref{tab:transformer} and ~\ref{tab:transformer_gdjst}.
Our proposed method outperforms in most cases.

\begin{table*}[t]
\centering
\caption{Quantitative results of motion data augmentation on the Transformer-based human motion prediction~\cite{DBLP:conf/iccvw/Martinez-Gonzalez21}.}
\begin{tabular}{lcc|lllllllll}
\hline
\multicolumn{3}{l|}{Methods} & \multicolumn{9}{l}{Prediction errors$\downarrow$[rad] on each action class \& timesteps [ms]} \\ \hline
{\color[HTML]{333333} } & {\color[HTML]{333333} } & {\color[HTML]{333333} } & \multicolumn{3}{c|}{{\color[HTML]{333333} punch}} & \multicolumn{3}{c|}{{\color[HTML]{333333} kick}} & \multicolumn{3}{c}{{\color[HTML]{333333} walk}} \\
\multirow{-2}{*}{{\color[HTML]{333333} }} & \multirow{-2}{*}{{\color[HTML]{333333} \begin{tabular}[c]{@{}c@{}}physical\\ correction\end{tabular}}} & \multirow{-2}{*}{{\color[HTML]{333333} \begin{tabular}[c]{@{}c@{}}motion\\ debiasing\end{tabular}}} & {\color[HTML]{333333} 100} & {\color[HTML]{333333} 200} & \multicolumn{1}{l|}{{\color[HTML]{333333} 400}} & {\color[HTML]{333333} 100} & {\color[HTML]{333333} 200} & \multicolumn{1}{l|}{{\color[HTML]{333333} 400}} & {\color[HTML]{333333} 100} & {\color[HTML]{333333} 200} & {\color[HTML]{333333} 400} \\ \hline
{\color[HTML]{333333} No aug} & {\color[HTML]{333333} -} & {\color[HTML]{333333} -} & {\color[HTML]{333333} 0.61} & {\color[HTML]{333333} 1.66} & \multicolumn{1}{l|}{{\color[HTML]{333333} 2.55}} & {\color[HTML]{333333} 0.5} & {\color[HTML]{333333} 1.48} & \multicolumn{1}{l|}{{\color[HTML]{333333} 2.20}} & {\color[HTML]{333333} 0.27} & {\color[HTML]{333333} 0.93} & {\color[HTML]{333333} 1.65} \\
{\color[HTML]{333333} Noise} & {\color[HTML]{333333} -} & {\color[HTML]{333333} -} & {\color[HTML]{333333} 0.60} & {\color[HTML]{333333} 1.65} & \multicolumn{1}{l|}{{\color[HTML]{333333} 2.53}} & {\color[HTML]{333333} 0.50} & {\color[HTML]{333333} 1.47} & \multicolumn{1}{l|}{{\color[HTML]{333333} 2.20}} & {\color[HTML]{333333} 0.27} & {\color[HTML]{333333} 0.88} & {\color[HTML]{333333} 1.55} \\ \hline
{\color[HTML]{333333} VAE} & \multicolumn{1}{l}{{\color[HTML]{333333} }} & \multicolumn{1}{l|}{{\color[HTML]{333333} }} & {\color[HTML]{333333} \textbf{0.47}} & {\color[HTML]{333333} \textbf{1.06}} & \multicolumn{1}{l|}{{\color[HTML]{333333} \textbf{1.51}}} & {\color[HTML]{333333} 0.40} & {\color[HTML]{333333} 1.01} & \multicolumn{1}{l|}{{\color[HTML]{333333} 1.42}} & {\color[HTML]{333333} 0.22} & {\color[HTML]{333333} 0.58} & {\color[HTML]{333333} 0.89} \\
{\color[HTML]{333333} IK} & \multicolumn{1}{l}{{\color[HTML]{333333} }} & \multicolumn{1}{l|}{{\color[HTML]{333333} }} & {\color[HTML]{333333} 0.49} & {\color[HTML]{333333} 1.14} & \multicolumn{1}{l|}{{\color[HTML]{333333} 1.58}} & {\color[HTML]{333333} 0.43} & {\color[HTML]{333333} 1.15} & \multicolumn{1}{l|}{{\color[HTML]{333333} 1.67}} & {\color[HTML]{333333} 0.30} & {\color[HTML]{333333} 0.86} & {\color[HTML]{333333} 1.31} \\
{\color[HTML]{333333} VAE \& IK} & \multicolumn{1}{l}{{\color[HTML]{333333} }} & \multicolumn{1}{l|}{{\color[HTML]{333333} }} & {\color[HTML]{333333} \textbf{0.47}} & {\color[HTML]{333333} \textbf{1.06}} & \multicolumn{1}{l|}{{\color[HTML]{333333} 1.49}} & {\color[HTML]{333333} 0.41} & {\color[HTML]{333333} 1.00} & \multicolumn{1}{l|}{{\color[HTML]{333333} 1.38}} & {\color[HTML]{333333} 0.26} & {\color[HTML]{333333} 0.79} & {\color[HTML]{333333} 1.25} \\ \hline
{\color[HTML]{333333} VAE} & {\color[HTML]{333333} \checkmark} & {\color[HTML]{333333} } & {\color[HTML]{333333} 0.51} & {\color[HTML]{333333} 1.22} & \multicolumn{1}{l|}{{\color[HTML]{333333} 1.71}} & {\color[HTML]{333333} 0.44} & {\color[HTML]{333333} 1.15} & \multicolumn{1}{l|}{{\color[HTML]{333333} 1.64}} & {\color[HTML]{333333} 0.22} & {\color[HTML]{333333} 0.58} & {\color[HTML]{333333} 0.92} \\
{\color[HTML]{333333} IK} & {\color[HTML]{333333} \checkmark} & {\color[HTML]{333333} } & {\color[HTML]{333333} 0.51} & {\color[HTML]{333333} 1.10} & \multicolumn{1}{l|}{{\color[HTML]{333333} 1.56}} & {\color[HTML]{333333} 0.45} & {\color[HTML]{333333} 1.17} & \multicolumn{1}{l|}{{\color[HTML]{333333} 1.70}} & {\color[HTML]{333333} 0.28} & {\color[HTML]{333333} 0.79} & {\color[HTML]{333333} 1.24} \\
{\color[HTML]{333333} VAE \& IK} & {\color[HTML]{333333} \checkmark} & {\color[HTML]{333333} } & {\color[HTML]{333333} 0.55} & {\color[HTML]{333333} 1.25} & \multicolumn{1}{l|}{{\color[HTML]{333333} 1.83}} & {\color[HTML]{333333} 0.46} & {\color[HTML]{333333} 1.15} & \multicolumn{1}{l|}{{\color[HTML]{333333} 1.68}} & {\color[HTML]{333333} 0.22} & {\color[HTML]{333333} 0.59} & {\color[HTML]{333333} 0.96} \\ \hline
{\color[HTML]{333333} VAE} & {\color[HTML]{333333} \checkmark} & {\color[HTML]{333333} \checkmark} & {\color[HTML]{333333} 0.52} & {\color[HTML]{333333} 1.18} & \multicolumn{1}{l|}{{\color[HTML]{333333} 1.69}} & {\color[HTML]{333333} 0.42} & {\color[HTML]{333333} 1.04} & \multicolumn{1}{l|}{{\color[HTML]{333333} 1.48}} & {\color[HTML]{333333} 0.23} & {\color[HTML]{333333} 0.60} & {\color[HTML]{333333} 0.97} \\
{\color[HTML]{333333} IK} & {\color[HTML]{333333} \checkmark} & {\color[HTML]{333333} \checkmark} & {\color[HTML]{333333} \textbf{0.47}} & {\color[HTML]{333333} 1.12} & \multicolumn{1}{l|}{{\color[HTML]{333333} 1.63}} & {\color[HTML]{333333} 0.39} & {\color[HTML]{333333} 0.86} & \multicolumn{1}{l|}{{\color[HTML]{333333} 1.18}} & {\color[HTML]{333333} 0.21} & {\color[HTML]{333333} 0.59} & {\color[HTML]{333333} 0.95} \\
{\color[HTML]{333333} VAE \& IK} & {\color[HTML]{333333} \checkmark} & {\color[HTML]{333333} \checkmark} & {\color[HTML]{333333} 0.51} & {\color[HTML]{333333} 1.18} & \multicolumn{1}{l|}{{\color[HTML]{333333} 1.73}} & {\color[HTML]{333333} \textbf{0.38}} & {\color[HTML]{333333} \textbf{0.84}} & \multicolumn{1}{l|}{{\color[HTML]{333333} \textbf{1.15}}} & {\color[HTML]{333333} \textbf{0.19}} & {\color[HTML]{333333} \textbf{0.48}} & {\color[HTML]{333333} \textbf{0.77}} \\ \hline
\end{tabular}
\label{tab:transformer}
\end{table*}

\begin{table*}[t]
\centering
\caption{Quantitative results of motion data augmentation on the Transformer-based human motion prediction~\cite{DBLP:conf/iccvw/Martinez-Gonzalez21} on \textit{grab}, \textit{deposit}, \textit{jog}, \textit{sneak}, \textit{throw}.}
\begin{tabular}{lcc|llllllllll}
\hline
\multicolumn{3}{l|}{Methods} & \multicolumn{10}{c}{action class \& timesteps [ms]} \\ \hline
\multicolumn{1}{c}{\multirow{2}{*}{}} & \multirow{2}{*}{\begin{tabular}[c]{@{}c@{}}physical\\ correction\end{tabular}} & \multirow{2}{*}{\begin{tabular}[c]{@{}c@{}}motion\\ debiasing\end{tabular}} & \multicolumn{2}{c|}{grab$\downarrow$} & \multicolumn{2}{c|}{deposit$\downarrow$} & \multicolumn{2}{c|}{jog$\downarrow$} & \multicolumn{2}{c|}{sneak$\downarrow$} & \multicolumn{2}{c}{throw$\downarrow$} \\
\multicolumn{1}{c}{} &  &  & 100 & \multicolumn{1}{l|}{400} & 100 & \multicolumn{1}{l|}{400} & 100 & \multicolumn{1}{l|}{400} & \multicolumn{1}{c}{100} & \multicolumn{1}{l|}{400} & \multicolumn{1}{c}{100} & 400 \\ \hline
No aug & - & - & 0.26 & \multicolumn{1}{l|}{1.83} & 0.25 & \multicolumn{1}{l|}{\textbf{1.80}} & 0.26 & \multicolumn{1}{l|}{1.11} & 0.14 & \multicolumn{1}{l|}{1.04} & 0.48 & 2.36 \\
Noise & - & - & 0.26 & \multicolumn{1}{l|}{1.83} & 0.25 & \multicolumn{1}{l|}{1.80} & 0.27 & \multicolumn{1}{l|}{1.13} & 0.14 & \multicolumn{1}{l|}{1.01} & 0.48 & 2.36 \\ \hline
VAE & \multicolumn{1}{l}{} & \multicolumn{1}{l|}{} & 0.24 & \multicolumn{1}{l|}{1.20} & \textbf{0.23} & \multicolumn{1}{l|}{1.34} & 0.26 & \multicolumn{1}{l|}{1.13} & 0.13 & \multicolumn{1}{l|}{0.80} & 0.46 & 2.02 \\
IK & \multicolumn{1}{l}{} & \multicolumn{1}{l|}{} & 0.23 & \multicolumn{1}{l|}{1.25} & \textbf{0.23} & \multicolumn{1}{l|}{1.32} & 0.25 & \multicolumn{1}{l|}{1.02} & 0.11 & \multicolumn{1}{l|}{0.71} & 0.47 & 2.00 \\
VAE \& IK & \multicolumn{1}{l}{} & \multicolumn{1}{l|}{} & 0.32 & \multicolumn{1}{l|}{1.32} & 0.26 & \multicolumn{1}{l|}{1.41} & 0.25 & \multicolumn{1}{l|}{1.02} & 0.11 & \multicolumn{1}{l|}{0.69} & 0.56 & 2.49 \\ \hline
VAE & \checkmark & \multicolumn{1}{l|}{} & 0.23 & \multicolumn{1}{l|}{1.29} & 0.24 & \multicolumn{1}{l|}{1.47} & 0.26 & \multicolumn{1}{l|}{1.09} & 0.12 & \multicolumn{1}{l|}{0.72} & 0.47 & 2.15 \\
IK & \checkmark & \multicolumn{1}{l|}{} & 0.23 & \multicolumn{1}{l|}{1.21} & 0.24 & \multicolumn{1}{l|}{1.34} & 0.25 & \multicolumn{1}{l|}{1.07} & 0.11 & \multicolumn{1}{l|}{0.69} & \textbf{0.41} & \textbf{1.56} \\
VAE \& IK & \checkmark &  & 0.23 & \multicolumn{1}{l|}{1.24} & 0.29 & \multicolumn{1}{l|}{1.75} & 0.25 & \multicolumn{1}{l|}{1.08} & 0.12 & \multicolumn{1}{l|}{0.72} & 0.44 & 1.73 \\ \hline
VAE & \checkmark & \checkmark & 0.22 & \multicolumn{1}{l|}{1.20} & \textbf{0.23} & \multicolumn{1}{l|}{1.43} & 0.25 & \multicolumn{1}{l|}{1.00} & 0.11 & \multicolumn{1}{l|}{0.66} & 0.45 & 1.96 \\
IK & \checkmark & \checkmark & \textbf{0.20} & \multicolumn{1}{l|}{\textbf{1.04}} & \textbf{0.23} & \multicolumn{1}{l|}{\textbf{1.29}} & \textbf{0.23} & \multicolumn{1}{l|}{0.85} & \textbf{0.10} & \multicolumn{1}{l|}{\textbf{0.60}} & 0.43 & 1.73 \\
VAE \& IK & \checkmark & \checkmark & 0.26 & \multicolumn{1}{l|}{1.11} & 0.25 & \multicolumn{1}{l|}{1.53} & \textbf{0.23} & \multicolumn{1}{l|}{\textbf{0.83}} & \textbf{0.10} & \multicolumn{1}{l|}{\textbf{0.60}} & 0.45 & 2.02 \\ \hline
\end{tabular}
\label{tab:transformer_gdjst}
\end{table*}

\section{Performance Comparison on Augmentation by DeepMimic}
We evaluated the augmentation by DeepMimic on human motion prediction.
All experimental settings, including the sampling of targets, follow Sec 4.3.
The result is shown in Table~\ref{tab:deepmimic_aug}.
DeepMimic has limited performance compared to our approach.

\begin{table}[t]
\centering
\caption{Performance comparison of DeepMimic augmentation on \textit{kick}.}
\begin{tabular}{l|lll}
\hline
{\color[HTML]{333333} Prediction errors on action class$\downarrow$} & \multicolumn{3}{c}{{\color[HTML]{333333} kick}} \\
{\color[HTML]{333333} timesteps[ms]} & {\color[HTML]{333333} 100} & {\color[HTML]{333333} 200} & {\color[HTML]{333333} 400} \\ \hline
GCN+No Aug & 1.08 & 1.68 & 2.26 \\
GCN+ours & \textbf{0.52} & \textbf{1.23} & \textbf{1.74} \\
GCN+ours(w/o residual force) & 1.07 & 1.65 & 2.08 \\
GCN+DeepMimic augmentation & 1.13 & 1.72 & 2.24 \\ \hline
\end{tabular}
\label{tab:deepmimic_aug}
\end{table}

\end{document}